\documentclass[10pt,twocolumn,letterpaper]{article}

\usepackage{cvpr}
\usepackage{times}
\usepackage{epsfig}
\usepackage{graphicx}
\usepackage{amsmath}
\usepackage{amssymb}
\usepackage{float}
\usepackage{subcaption}
\usepackage{multirow}
\usepackage{enumitem,kantlipsum}
\usepackage{xcolor}
\usepackage{booktabs}
\usepackage{multirow} 
\makeatletter
\robustify\@latex@warning@no@line
\makeatother
\usepackage{authblk}
\restylefloat{figure}     
\makeatletter
\renewcommand\AB@affilsepx{, \protect\Affilfont}
\makeatother

\newcommand\blfootnote[1]{%
  \begingroup
  \renewcommand\thefootnote{}\footnote{#1}%
  \addtocounter{footnote}{-1}%
  \endgroup
}

\def\etal{\emph{et al}\onedot}

\usepackage[pagebackref=true,breaklinks=true,letterpaper=true,colorlinks,bookmarks=false]{hyperref}

\begin{document}




\title{OutfitTransformer: Learning Outfit Representations \\ for Fashion Recommendation}
\author[1,2]{Rohan Sarkar}
\author[2]{Navaneeth Bodla}
\author[2]{Mariya I. Vasileva}
\author[2]{Yen-Liang Lin}
\author[2]{Anurag Beniwal}
\author[2]{Alan Lu}
\author[2]{Gerard Medioni}
\affil[1]{Purdue University, West Lafayette}
\affil[2]{Amazon}




\maketitle
\begin{abstract}
\vspace*{-.5em}
Learning an effective outfit-level representation is critical for predicting the compatibility of items in an outfit, and retrieving complementary items for a partial outfit. We present a framework, OutfitTransformer, that uses the proposed task-specific tokens and leverages the self-attention mechanism to learn effective outfit-level representations encoding the compatibility relationships between all items in the entire outfit for addressing both compatibility prediction and complementary item retrieval tasks.
%
%
For compatibility prediction, we design an outfit token to capture a global outfit representation and train the framework using a classification loss. For complementary item retrieval, we design a target item token that additionally takes the target item specification (in the form of a category or text description) into consideration. We train our framework using a proposed set-wise outfit ranking loss to generate a target item embedding given an outfit, and a target item specification as inputs. The generated target item embedding is then used to retrieve compatible items that match the rest of the outfit. Additionally, we adopt a pre-training approach and a curriculum learning strategy to improve retrieval performance. Since our framework learns at an outfit-level, it allows us to learn a single embedding capturing higher-order relations among multiple items in the outfit more effectively than pairwise methods.
Experiments demonstrate that our approach outperforms state-of-the-art methods on compatibility prediction, fill-in-the-blank, and complementary item retrieval tasks.
We further validate the quality of our retrieval results with a user study.
\end{abstract}

\section{Introduction}
\blfootnote{Emails: sarkarr@purdue.edu, navaneeth.bodla@getcruise.com, \{vamariy, yenliang, beanurag, alalu, medioni\}@amazon.com}
Two main tasks for a fashion outfit recommendation system are fashion {\em compatibility prediction} and large-scale {\em complementary item retrieval}. For compatibility prediction (CP), the task is to determine whether a set of fashion items in an outfit go well together. For complementary item retrieval (CIR), the task is to complete a partial outfit by finding a compatible item from a large database.  

\begin{figure}[t]
    \centering
    \includegraphics[width=0.5\textwidth]{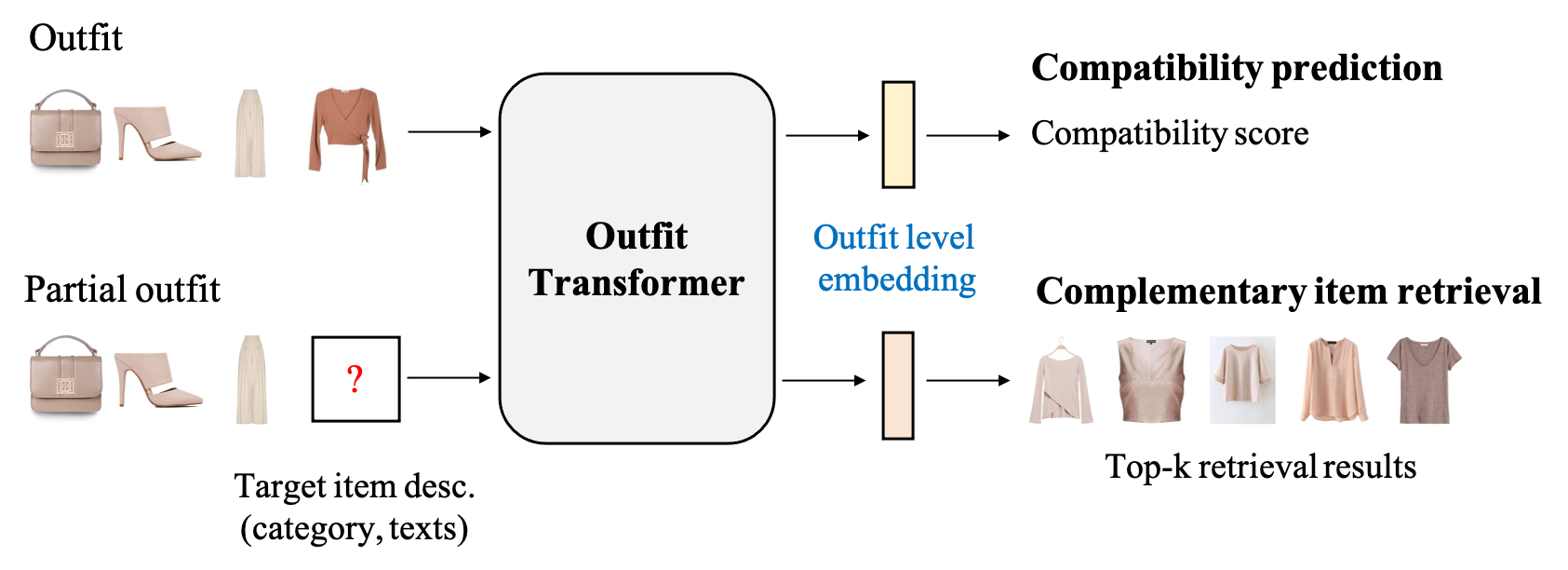}
    \caption{The proposed framework, OutfitTransformer, learns an outfit-level representation for a set of outfit items to address the CP and CIR tasks. For CIR, it learns a single embedding encoding overall compatibility of the partial outfit, and  a target item description that is used to retrieve compatible items cohesively matching the entire outfit using KNN search. For CP, our framework learns an outfit-level representation capturing overall outfit compatibility to predict a compatibility score.}
    \label{fig:fig1}
\end{figure}

Given an outfit, we want to predict how well its constituent items go together.
Also, given a partial outfit with different items (such as a bag, shoes, and pants) and a target item description (e.g., ``top''), we want to retrieve compatible items to complete the outfit. Figure \ref{fig:fig1} illustrates our proposed method.


Prior work such as \cite{Veit2015,Veit2017,typeaware2018,scenet2019,Yang2019TransNFCMTN} addresses the pairwise item-level compatibility problem and achieves state-of-the-art results but does not explicitly model outfit-level compatibility. 
Some methods optimize for compatibility at an outfit-level \cite{hanACMMM2017,Cucurull_2019_CVPR,Hsiao_2017_ICCV,Hsiao_2018_CVPR,POG2019}. However, these approaches are mainly designed for classification tasks: compatibility prediction and fill-in-the-blank (FITB) but they do not address the large-scale CIR task.
%
CSA-Net \cite{compretrieval2020} proposes a method for large-scale CIR, but it does not 
learn an outfit-level representation that can explicitly capture compatibility of
a target item to the outfit as a whole.
It searches compatible items for each item in the outfit
at a paired-category level (e.g., top to shoe, bottom to shoe) and fuses the ranking scores for different query items to
obtain the final rankings.

Instead, our idea is to learn an outfit-level representation for both compatibility prediction and large-scale retrieval of complementary items. 
Here, we investigate a transformer-based architecture to learn the outfit-level embeddings for the CIR task, as this architecture shows better outfit compatibility prediction performance than the Bi-LSTM \cite{hanACMMM2017} and GCN-based \cite{Cucurull_2019_CVPR} architectures (cf. Table \ref{tbl:comp}).  
Specifically for CIR, the target item should cohesively match all the existing items in an outfit.
Using outfit-level representations can more effectively capture complex feature correlations among multiple items in the outfit, as opposed to considering pairs of items at a time.
Additionally, users may specify their preference for the target item in the form of a target category (e.g., top) or the text descriptions (e.g., full sleeve shirt with floral design), and our system is able to retrieve the complementary items that match it. 
%
%
We design our framework to learn a target item embedding that operates at the outfit-level and encodes both the overall compatibility of a partial outfit, and the target item specification. 
We pose this as a set-to-item compatibility learning problem where we model the outfit as a set of items, and extract a single target item embedding to search for complementary items.

The outfit representation learnt for both tasks is invariant to the order of the items, i.e., permutation of the order of the outfit items should generate the same representation. 
The transformer is a suitable choice for our framework because it captures the higher-order relationships (beyond pairwise) between all constituent items in the outfit and is able to take unordered items as input. 

For the task of CP, we train the OutfitTransformer with a classification loss and design an outfit token to capture a global outfit representation that encodes the compatibility relationships among all the items in the outfit. 
For CIR, we design a target item token that encodes both the compatibility of the partial outfit and a target item description to generate the embedding of the target item.  This embedding is used to retrieve compatible items from a database. 
We train our framework using a proposed set-wise outfit ranking loss, which encourages compatible items to be embedded closer to the overall representation of a set of outfit items.
Our design allows extraction of a single target item embedding, which enables large-scale indexing and retrieval. \footnote{The framework needs to be designed in a way such that it allows individual item embedding extraction (should not depend on the query image during indexing like SCE-Net \cite{scenet2019}) to support large-scale indexing for KNN search. (cf. Sec \ref{sec:indexing_and_retrieval}).}


Directly training on the retrieval task leads to poor performance, since the network does not have any prior knowledge regarding compatibility of the partial outfit. To alleviate this problem, we facilitate OutfitTransformer to learn compatibility relationships by pre-training it on the CP task. We find that this improves retrieval performance significantly (cf. Table \ref{tbl:abl_retr}(a)). 
%
In addition, we propose a curriculum learning strategy to 
hierarchically sample more informative negative examples
which further boosts retrieval performance (cf. Table \ref{tbl:abl_retr}(b)).

We evaluate our method on the public Polyvore Outfits dataset \cite{typeaware2018}.
Experimental results show that our approach outperforms state-of-the-art techniques in compatibility prediction, fill-in-the-blank (FITB), and complementary item retrieval tasks. In Section \ref{sec:results}, we demonstrate that our framework can retrieve complementary items based on the target item category or description. In addition, we validate our results for the CIR task using Amazon Mechanical Turk.

In summary, our main technical contributions are:
\begin{itemize}





\item We propose a new framework, OutfitTransformer, that effectively learns outfit-level representations, which is shown experimentally to outperform state-of-the-art methods on both compatibility prediction (CP) and complementary item retrieval (CIR) tasks.

\item We propose task-specific tokens to support both CP and CIR. For CP, the outfit token is designed to capture a global outfit representation. For CIR, the target item token additionally takes the target specification (in the form of category or free-form text) into account. 

\item Our framework learns a single embedding that enables large-scale indexing and retrieval for complementary items, and has smaller indexing size than previous approaches
(\cite{compretrieval2020,typeaware2018}) which use subspace embeddings.

\item We provide in-depth analysis of different design choices (pre-training and curriculum learning) to improve retrieval performance.

%
%

\end{itemize}

\section{Related Work}
\label{sec:related_work}
\noindent

\textbf{Outfit Compatibility Prediction.} Prior work on fashion outfit compatibility
often considers pairwise item comparisons to predict the final compatibility score~\cite{Veit2015,Veit2017,typeaware2018,scenet2019,Yang2019TransNFCMTN}.
However, these algorithms usually perform aggregation of item-level scores to compute the entire outift compatibility score. 
To add global constraints in scoring outfit compatibility, a number of methods \cite{hanACMMM2017,Cucurull_2019_CVPR,Hsiao_2018_CVPR,POG2019} aggregate inputs from all constituent items. 
Han \etal~\cite{hanACMMM2017} use a bidirectional LSTM to model outfit composition as a sequential process, considering outfits as ordered sequences. However, outfit compatibility should be invariant to the order of items in an outfit. 
Some recent approaches~\cite{Cucurull_2019_CVPR,NGNN2019} use a graph convolutional network (GCN) for outfit compatibility prediction. 
Cucurull \etal~\cite{Cucurull_2019_CVPR} train a graph neural network that generates embeddings conditioned on the representations of neighboring nodes.
They use GCN to model the node relationship and predict the outfit compatibility but require large neighbor information for best performance, which is impractical for new items as mentioned in \cite{compretrieval2020}. 
Cui \etal~\cite{NGNN2019} model an outfit as a graph, where each node represents a category and each edge represents interaction between two categories. 
%
Chen~\etal~\cite{POG2019} (\emph{hrta.} POG) propose a transformer-based encoder-decoder architecture for generating compatible outfits that are specifically designed for personalization based on historical clicks data.
All these approaches~\cite{hanACMMM2017,Cucurull_2019_CVPR,NGNN2019,POG2019} are mainly designed for classification tasks (CP and FITB \cite{hanACMMM2017}) but do not address large-scale CIR.


\noindent
\textbf{Outfit Complementary Item Retrieval.} 
Although the compatibility prediction score in previous methods can be used for ranking items, it is impractical to do so in a large-scale setting. The framework must support indexing to avoid linearly scanning the entire database. 
Lin \etal~\cite{compretrieval2020} propose a framework to retrieve complementary items that match the entire outfit. Their method retrieves items by considering compatibility between the target item and every
item in an outfit in a pairwise manner, and then aggregating the scores. However, they do not consider the outfit as a whole and only use attention at a paired-category level. In contrast, we use a transformer model to capture interactions between all the items in an outfit to learn a global outfit representation. Aso, our method has a much smaller indexing size than \cite{compretrieval2020} which is important for practical applications (cf. Section \ref{sec:indexing_and_retrieval}).
Lorbert~\etal~\cite{lorbert2021} use a single layer self-attention based framework for outfit generation but do not explicitly model compatibility. 
%
%

\noindent 
\textbf{Attention-Based Methods and Vision Transformers.} There has been considerable research using transformers in a wide variety of computer vision tasks \cite{ObjectDetectionTransform,segformer,vaswani-vision-traform,super-resolution-transformer}. Vision transformers \cite{vit2021} and related models \cite{vision-transformer-distillation,focal-transformer,CrossViT2021} decompose each image into an ordered sequence of smaller patches to learn image representation of a single image for different tasks. On the contrary, we model outfits as an unordered set of different item images and use a transformer-based architecture to learn a global outfit representation that captures overall outfit compatibility. As discussed earlier in this section, attention mechanisms \cite{compretrieval2020,Meet2021,POG2019,lorbert2021} have also been used in fashion recommendation systems. \cite{compretrieval2020,Meet2021} use attention to understand complementary relationships in a pairwise manner. In contrast, we use a transformer model to learn interactions between all the items in an outfit, which attends to higher-order compatibility relationships~\cite{SetT2018} beyond pairwise~\cite{typeaware2018,compretrieval2020,scenet2019}. 
%
Both POG \cite{POG2019} and \cite{lorbert2021} use pretrained ImageNet embeddings, while we learn fashion-specific features by training in an end-to-end manner. 

\noindent \textbf{Distance Metric Learning:} CIR is a fundamentally different task from visual similarity search because the complementary item is from a different category and is visually dissimilar from the other items in the outfit. We specifically design a negative sampling strategy for CIR where we sample negatives from the same category as positive samples, unlike \cite{facenet2015,Song2016DeepML}.  
Also, in contrast to \cite{scenet2019,typeaware2018,compretrieval2020}, which consider pair-wise compatibility between target items and each individual item in the outfit, we propose a set-wise outfit ranking loss that compares target items with a single embedding for the entire outfit.



\section{Proposed Approach}
\begin{figure*}[t]
   \centering
   \begin{minipage}[c]{0.45\textwidth}
   \includegraphics[width=\textwidth]{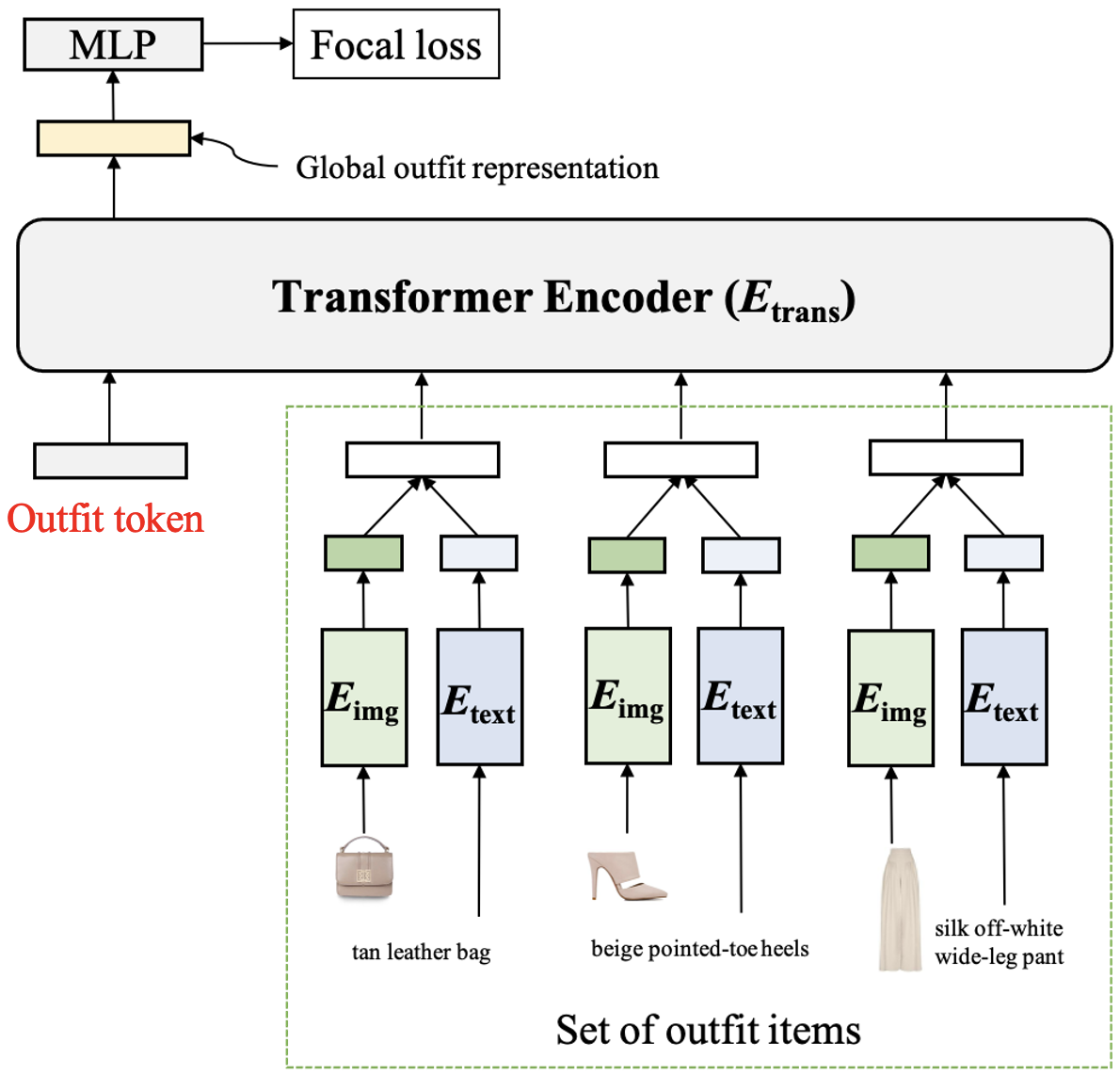}
   \subcaption{\hspace{0.5in} \small Compatibility prediction}
   \end{minipage}
   \hfill \vline \hfill 
   \begin{minipage}[c]{0.49\textwidth}
   \includegraphics[width=\textwidth]{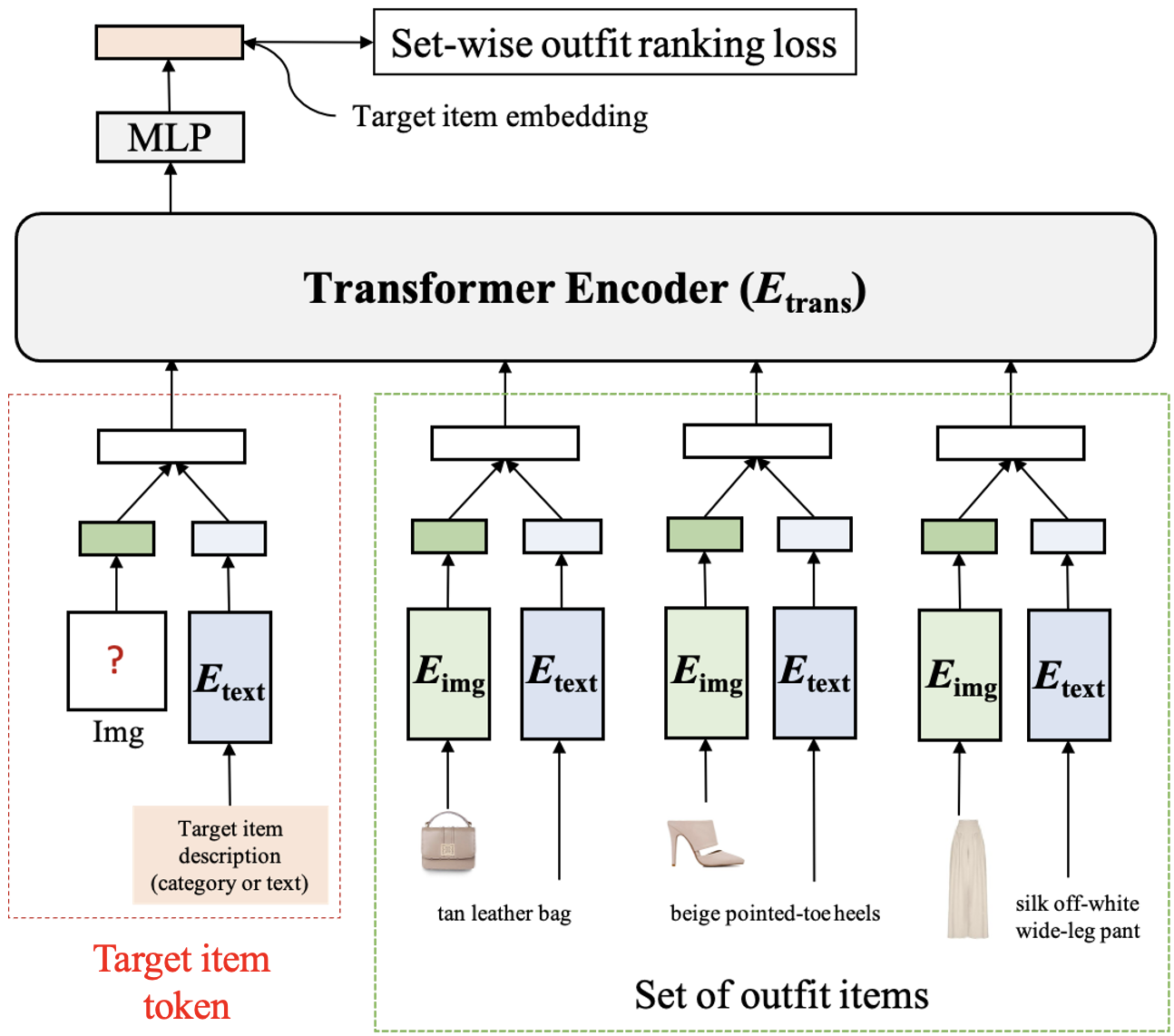}
   \subcaption{\small Complementary item retrieval}
   \end{minipage}
   \caption{System overview of our framework for compatibility prediction and complementary item retrieval. We model outfits as an unordered set of items. We use an image encoder ($E_{\mathrm{img}}$) and a text encoder ($E_{\mathrm{text}}$) to extract the image and text features. (a) For compatibility prediction, we train the transformer encoder using a focal loss \cite{focal2020} and learn a global outfit representation to predict an outfit compatibility score. (b) For complementary item retrieval, given an outfit and a target item description, we train the transformer encoder to learn a target item embedding that can be used for retrieving compatible items to complete an outfit. We train the framework using the proposed set-wise outfit ranking loss in an end-to-end manner. The details of set-wise ranking loss are explained in Section \ref{sec:ocloss}.  
   \label{fig:Overview}}
\end{figure*}

Figure \ref{fig:Overview} illustrates the overview of our framework. 
Our framework takes as input each outfit's constituent item images and their text descriptions. 
%
%
%
For compatibility prediction, we train the transformer encoder to generate a global outfit representation that can capture  higher-order compatibility relationships between all items in the outfit beyond pairwise relationships.
This global outfit representation can then be used to predict an outfit compatibility score. Details are presented in Section \ref{sec:compat}. 
%
%
%

For complementary item retrieval, given a partial outfit and a target item description (e.g., product category or description), we train the transformer encoder to generate a target item embedding, which can be used to retrieve items that are compatible with the partial outfit and match the target item description. 
The framework is trained using a proposed ranking loss that enforces the target item embedding to move closer to the positive item and further apart from the negative items. The positive item matches the global style of the overall outfit, whereas the negatives are incompatible with the outfit (details in Section \ref{sec:retrieval}).

We investigate different training strategies to improve retrieval performance. We employ a pre-training strategy where we first train the model on the compatibility prediction task as mentioned in Section \ref{sec:pretrain}. We also adopt curriculum learning to select more informative negative samples in different training stages. 
The details are presented in the Section \ref{sec:ocloss}.

\subsection{Fashion Outfit Compatibility Prediction}
\label{sec:compat}
The compatibility prediction task predicts the compatibility of all the items in an outfit. Given an outfit $O=\left\{\left({\text{I}}_i,{\mathrm T}_i\right)\right\}_{i=1}^L$, where ${\mathrm I}_i$ is the image, ${\mathrm T}_i$ is the corresponding text description for an item $i$. We learn a non-linear function that predicts a compatibility score in $[0, 1]$, where $1$ indicates perfect compatibility.

The item images and their text descriptions are fed into an image ($E_{\mathrm{img}}$) and text encoder ($E_{\mathrm{text}}$) respectively to extract the image and text feature vectors (see Section \ref{sec:training} for the details about the image and text encoder architecture). We concatenate the extracted image and text feature vectors to generate an item feature vector $u_i= 
E_{\mathrm{img}}\left({\text{I}}_i\right)\parallel E_\text{text}\left({\mathrm T}_i\right)$, where $\parallel$ denotes a concatenation operation. 
The set ${F}=\left\{u_i\right\}_{i=1}^L$ represents the feature vectors of all outfit items.

Since the goal of vision transformers \cite{vit2021} is to produce a classification score for an image, a classifier token is typically used to capture a single image representation from the input image patches. In contrast, we introduce the outfit token whose state at the output of the transformer encoder serves as the global outfit representation. The goal of introducing this token is to learn a global outfit representation that captures compatibility relationships between items in the outfit using the self-attention mechanism. We model outfits as an unordered set of items as the overall outfit compatibility is invariant to the order of the items. Thus, positional encodings used in NLP~\cite{nlp2017} and ViT~\cite{vit2021} are not required for us. 

The outfit token (${\mathrm x}_\text{Outfit}$) is a learnable embedding that is prepended to the set of outfit feature vectors $F$ and fed into the transformer encoder ${\mathrm E}_\text{trans}$. The state of the outfit token at the output of the transformer encoder serves as the global outfit representation
which is subsequently fed into the MLP that predicts an overall outfit compatibility score:
\begin{equation}
\mathrm c=\mathrm{MLP}\left({\mathrm E}_\text{trans}\left({\mathrm x}_\text{Outfit},F\right)\right)
\end{equation}

Our framework ($E_{\mathrm{trans}},  E_{\mathrm{img}}, E_{\mathrm{text}}$) is trained end-to-end using focal loss \cite{focal2020}.

\subsection{Complementary Item Retrieval}
\label{sec:retrieval}
The complementary item retrieval task is to retrieve an item that is both compatible with the partial outfit and matches a specified item description to complete the outfit. Specifically, given a set of partial outfit items and a user provided target item specification, the goal is to generate a target item embedding that can be used to retrieve compatible items. Our framework is trained with a proposed set-wise outfit ranking loss which is discussed in detail in Section \ref{sec:ocloss}.

The target item token $\mathrm s$ (cf. Figure  \ref{fig:Overview} (b)) includes an item description $\mathrm T$ for the target item that we want to retrieve, and an empty image represented by $\mathrm x_{\mathrm{Img}}$. The target item token is defined as 
$
\mathrm s=\mathrm x_{\mathrm{Img}}^{}\parallel E_{\mathrm{text}}\left(\mathrm T\right)
$. 

The intuition behind designing the target item token in this manner is that, during inference, the target image is unknown but users can provide a description for the item they are searching for.
We simulate a similar setting when training the framework for the retrieval task. We introduce the target item token whose state at the output of the transformer encoder serves as the target item representation that explicitly takes into consideration both compatibility with the partial outfit, and the target item description.
Our framework is generic and the target item description can be provided in different forms such as category, text description, tags, etc.

The transformer encoder takes as input the set of feature vectors $F$ of the partial outfit, and the target item specification $\mathrm s$, which is subsequently fed into a MLP that generates the target item embedding. 
\begin{equation}
\mathrm t= \mathrm{MLP}\left(E_{\mathrm{trans}}\left(\mathrm s\mathit, F\right)\right)
\end{equation}

To learn this target item embedding, we train our framework with a proposed set-wise outfit ranking loss which is discussed in Section \ref{sec:ocloss}.

\subsubsection{Pre-training on Compatibility Prediction: }
\label{sec:pretrain}

We pre-train the framework on the compatibility prediction task and use the learned weights to initialize the transformer, image and text encoder for complementary item retrieval. This choice leads to a significant improvement 
for CIR (cf. Table \ref{tbl:abl_retr}(a)).

We conjecture that the reasons  might be: 1) the pre-trained transformer encoder captures compatibility relationships, which is helpful for encoding them into the target item embedding to retrieve compatible items, and 
 2) the image encoder captures fashion-specific features, which is used to extract better feature vectors for positive and negative samples in the set-wise outfit ranking loss.
%


\subsubsection{Set-wise Outfit Ranking Loss:} 
\label{sec:ocloss}
\begin{figure}[t]
   \centering
   \includegraphics[width=0.5\textwidth]{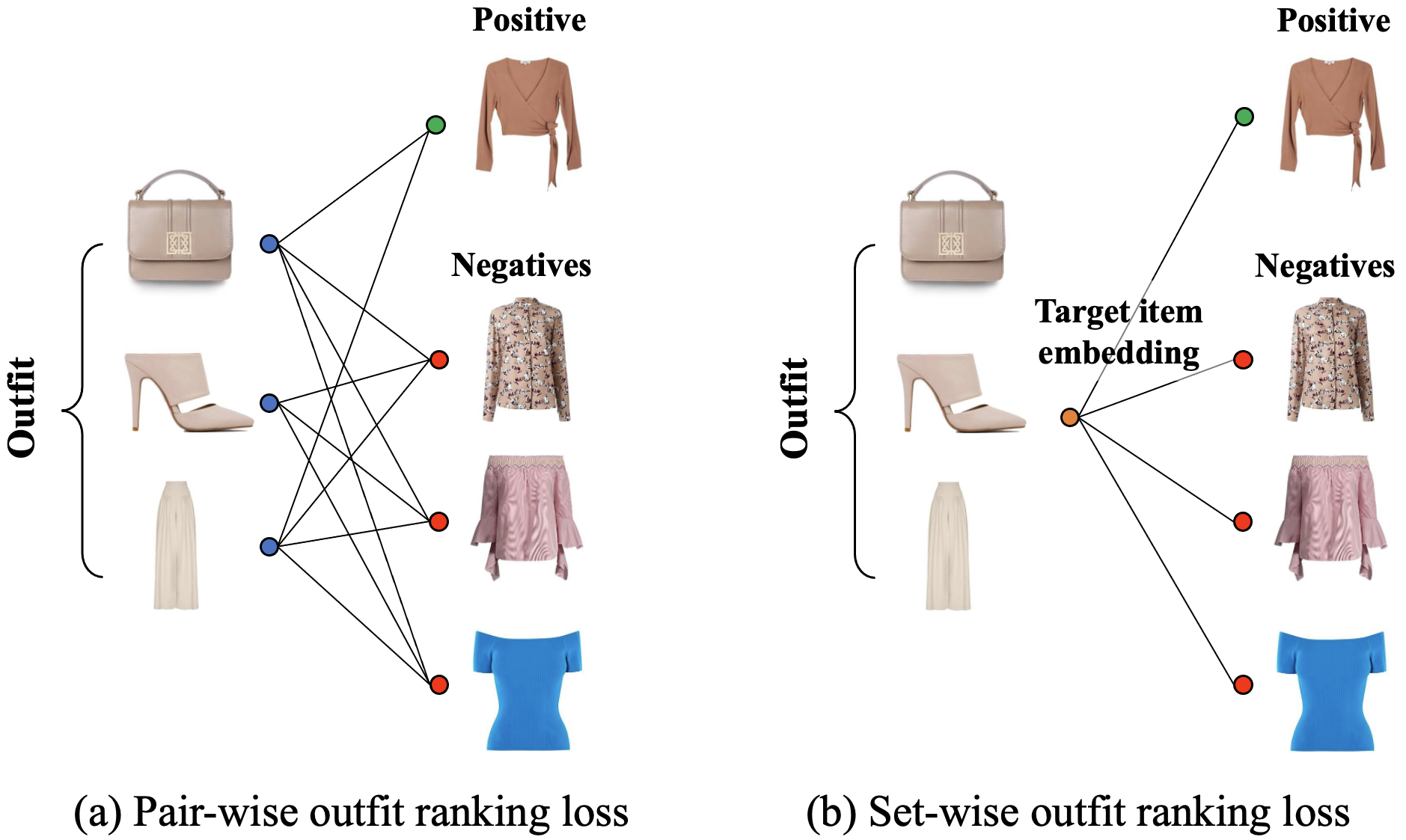}
   \caption{Comparison of pair-wise outfit ranking loss \cite{compretrieval2020} and our set-wise outfit ranking loss. Our framework generates a single target item embedding that captures the compatibility of the entire outfit and does not need pairwise computations with individual items in the outfit as in \cite{compretrieval2020}.}
    \label{fig:Loss}
\end{figure}

Previous approaches (\cite{typeaware2018}, \cite{scenet2019}) use a triplet loss to learn relationships only between a pair of items but do not consider the relationship between all items in the outfit. 
To address this, the outfit ranking loss \cite{compretrieval2020} is proposed which considers the pairwise compatibility of target items with all the items in the outfit, as shown in Figure \ref{fig:Loss} (a).
In contrast, our approach generates a single target item embedding $\mathrm t$ that already captures the compatibility relationships for a set of outfit items and hence does not require pairwise comparisons with individual outfit items, as shown in Figure \ref{fig:Loss} (b). This reduces the complexity of \cite{compretrieval2020} from $\mathcal{O}\left(LS\right)$ to $\mathcal{O}\left(S\right)$ during training, where $L$ denotes the outfit length and $S$ denotes the number of positive and negative samples.

%

In practice, only legitimate outfit samples are provided in the dataset, and there are no annotated negative samples. Given an outfit we randomly pick an item as positive and the remaining items as the partial outfit. Here, we investigate a curriculum learning approach to gradually increase the difficulty of negatives for training. Specifically, we train the model in two stages. In the first stage, we sample the negatives from the same high-level category as the positive item. Subsequently in the second stage, we sample harder negatives from more fine-grained categories. Note that since CIR is different from visual similarity search, our negatives are different from the conventional way of constructing triplets (e.g., \cite{facenet2015,Song2016DeepML}), where negatives are sampled from other classes. 

The set-wise outfit ranking loss is designed to optimize relative distances between samples such that the target item embedding moves closer to the positive embedding and farther apart from the negative embeddings. Note that we use the pre-trained image and text encoders (cf. Sec. \ref{sec:pretrain}) to extract the positive and negative embeddings. Because we have a single target item embedding that encodes the compatibility of the entire outfit, we can directly train our set-wise ranking loss using triplets without requiring pairwise computations as \cite{compretrieval2020}. The set-wise outfit ranking loss is defined as: 
\begin{equation}
L\left(\mathrm t,p,N\right)=\;L{\left(\mathrm t,p,N\right)}_{All}+\;L{\left(\mathrm t,p,N\right)}_{Hard}
\label{eq:setwise_outfit_ranking_loss}
\end{equation}
\noindent
\[
L{\left(\mathrm t,p,N\right)}_{All}=\frac1{\left|N\right|}\sum_{j=1}^{\left|N\right|}\left[d\left(\mathrm t,f^p\right)-d\left(\mathrm t,f_j^N\right)+m\right]_+
\]

\[
L{\left(\mathrm t,p,N\right)}_{Hard}={\left[d\left(\mathrm t,f^p\right)-\underset{j=1...\left|N\right|}{\text{min}}d\left(\mathrm t,f_j^N\right)+m\right]}_+
\]

\noindent 
where $[]_+$ is the hinge loss, $\mathrm t$ is the target item embedding, $f^p$ is the positive embedding, $f_j^N$ is the $j^{th}$ negative embedding from the pool of negative samples in $N$, and $m$ is the margin. 

The loss has two components as shown in Equation \eqref{eq:setwise_outfit_ranking_loss}. 
The first component $L{\left(\mathrm t,p,N\right)}_{All}$ considers all the sampled negatives for the outfit, while the second component $L{\left(\mathrm t,p,N\right)}_{Hard}$ considers the hard negative samples (e.g., \cite{hard2020,Song2016DeepML}). This  allows the model to learn discriminatory features to distinguish between items that might have very subtle differences between them. We empirically find that this loss formulation and the hard negative sampling strategy improves complementary item retrieval performance significantly (cf. Tables \ref{tbl:abl_retr}(b), \ref{tbl:abl_negsamp}). There are other sampling methods that could potentially be used (e.g., \cite{sampling2017}), but the investigation thereof is outside the scope of this paper. 

\subsubsection{Indexing and Retrieval for complementary items:}
\label{sec:indexing_and_retrieval}
%
Not all the methods for CP can support indexing for retrieval. For example, SCE-Net \cite{scenet2019} requires pairs of images as inputs, which does not allow single item embedding extraction for indexing. 
We design our framework in a way that allows extraction of individual item
feature vectors during indexing and generate a single item embedding during inference.
Based on our design, we can use off-the-shelf KNN search tools (e.g., \cite{bworld,faiss}) to perform indexing and retrieval, which makes the search very efficient even for a large database (e.g., with millions of items).
Specifically, during indexing, we use the trained image and text encoder to extract the item features. This does not depend on the query images unlike \cite{scenet2019}. 
During inference, given the partial outfit and a target item description, our framework generates a single target item embedding, which is then used to search for compatible items from the database using KNN search.

%
%

%

Our framework offers two advantages as compared to prior works. First, we require smaller indexing size compared to previous approaches that use subspace embeddings. For indexing, Type-aware \cite{typeaware2018} and CSA-Net \cite{compretrieval2020} generate multiple embeddings of each item for each of the target categories and therefore the indexing size
grows linearly with the number of categories. Because we are not learning subspaces, our approach is independent of the number of categories. Second, in \cite{compretrieval2020}, for each item in the outfit, a target category-specific embedding is extracted, which is used to retrieve compatible items from the database. This has to be repeated exhaustively for each item in the query outfit. In contrast, our framework can retrieve items in a single step regardless of outfit length. 

\begin{table}[b]
\centering\caption{Comparison of our model with state-of-the-art methods on the compatibility prediction task using the AUC metric \cite{hanACMMM2017}. The methods are evaluated on Polyvore-Outfits (where -D denotes the disjoint dataset). }
\begin{tabular}{@{}llcc@{}}
\toprule
Method & Features & PO-D & PO \\ \midrule
BiLSTM + VSE \cite{hanACMMM2017} & ResNet-18 + Text & 0.62 & 0.65 \\
GCN (k=0) \cite{coherent2019} & ResNet-18 & 0.67 & 0.68 \\
SiameseNet \cite{typeaware2018} & ResNet-18 & 0.81 & 0.81 \\
Type-Aware \cite{typeaware2018} & ResNet-18 + Text & 0.84 & 0.86 \\
SCE-Net \cite{scenet2019} & ResNet-18 + Text & - & 0.91 \\
CSA-Net \cite{compretrieval2020} & ResNet-18 & 0.87 & 0.91 \\
\midrule
OutfitTransformer (Ours) & ResNet-18 & 0.87 & 0.92 \\
OutfitTransformer (Ours) & ResNet-18 + Text & \textbf{0.88} & \textbf{0.93} \\ \bottomrule
\end{tabular}
\label{tbl:comp}
\end{table}
\subsection{Implementation Details}
\label{sec:training}
The image encoder uses a ResNet-18 initialized
with ImageNet pre-trained weights. The text encoder uses a pre-trained SentenceBERT \cite{SBERT2019}, on top of which we add a fc layer. During training, we finetune the weights of the image encoder and the fc layer of the text encoder. We extract a $64$-dimensional image and a $64$-dimensional text embedding and concatenate them to generate $128$-dimensional item embeddings before feeding them into the transformer encoder. We use a six-layer transformer encoder with $16$ heads. For the retrieval task, we set the margin $m$ for the set-wise outfit ranking loss as $2$ and sample 10 negatives for each outfit. We use a batch size of $50$ and optimize using ADAM with an initial learning rate of $1e-5$ and reducing the learning rate by half in steps of $10$.

\begin{table*}[t]
\centering
\caption{Comparison of our model with state-of-the-art methods on the FITB (using accuracy) and complementary item retrieval tasks (using recall@top-k). The methods are evaluated on Polyvore-Outfit Dataset (where -D denotes disjoint set).}
\begin{tabular}{@{}lcccccccc@{}}
\toprule
\multirow{2}{*}{Method} & \multicolumn{4}{c}{Polyvore Outfits-D} & \multicolumn{4}{c}{Polyvore Outfits} \\ \cmidrule(l){2-5} \cmidrule(l){6-9}
 & \multicolumn{1}{l}{FITB} & \multicolumn{1}{l}{R@10} & \multicolumn{1}{l}{R@30} & \multicolumn{1}{l}{R@50} & \multicolumn{1}{l}{FITB} & \multicolumn{1}{l}{R@10} & \multicolumn{1}{l}{R@30} & \multicolumn{1}{l}{R@50} \\ \midrule
Type-Aware \cite{typeaware2018} & 55.65 & 3.66 & 8.26 & 11.98 & 57.83 & 3.50 & 8.56 & 12.66 \\
SCE-Net Average \cite{scenet2019} & 53.67 & 4.41 & 9.85 & 13.87 & 59.07 & 5.10 & 11.20 & 15.93 \\
CSA-Net \cite{compretrieval2020} & 59.26 & 5.93 & \textbf{12.31} & \textbf{17.85} & 63.73 & 8.27 & 15.67 & 20.91 \\
\midrule
OutfitTransformer (Ours) & \textbf{59.48} & \textbf{6.53} & 12.12 & 16.64 & \textbf{67.10} & \textbf{9.58} & \textbf{17.96} & \textbf{21.98} \\ \bottomrule
\end{tabular}
\label{tbl:retr}
\end{table*}
\section{Experiments}
\label{sec:results}
We compare our proposed approach with the state-of-the-art baselines such as Bi-LSTM \cite{hanACMMM2017}, GCN \cite{Cucurull_2019_CVPR}, SiameseNet \cite{Veit2017}, Type-aware \cite{typeaware2018} , SCE-Net \cite{scenet2019} and CSA-Net \cite{compretrieval2020} on the Polyvore Outfits dataset \cite{typeaware2018}. For evaluation, we compare our method with these state-of-the-art baselines on three different tasks: 
\begin{enumerate}[wide, labelwidth=!, labelindent=0pt]
    \item {\em Outfit Compatibility Prediction (CP) task} that predicts the compatibility of items in an outfit. 
    \item {\em Fill in the Blank (FITB) task} that selects the most compatible item for an incomplete outfit given a set of candidate choices (e.g., 4 candidates).
    \item {\em Outfit Complementary Item Retrieval (CIR) task} that retrieves complementary items from the database for a target category given an incomplete outfit. 
\end{enumerate}

The Polyvore Outfits dataset \cite{typeaware2018} has two sets, the disjoint and non-disjoint sets. In the disjoint set, the training split items (and outfits) do not overlap with the validation and test splits. In the non-disjoint set, the training split items can overlap with those of validation and test splits, but outfits do not overlap. The non-disjoint set contains 53306 training and 10000 test outfits, while the disjoint set comprises of 16995 training and 15154 test outfits.

For the standard compatibility prediction and FITB tasks, we evaluate our model on the Polyvore Outfits dataset. Since the Polyvore Outfits dataset does not provide the annotations for the complementary item retrieval task, we adopt a modified version of the Polyvore Outfits dataset proposed by CSA-Net \cite{compretrieval2020}.

\subsection{Outfit Compatibility Prediction (CP)}

The goal of this task is to measure the compatibility of an outfit. 
Our compatibility model in Figure \ref{fig:Overview} (a) predicts a score that indicates the compatibility of the overall outfit. 
We compare the performance with the state-of-the-art methods in Table \ref{tbl:comp} by using the standard metric AUC \cite{hanACMMM2017}, which measures the area under the receiver operating characteristic curve.\footnote{~The authors do not report the performance of SCE-Net \cite{scenet2019} on the disjoint dataset.} While Bi-LSTM models outfits as a sequence of items, SiameseNet, Type-Aware, CSA-Net and SCE-Net
learn pairwise compatibility of items and aggregates the pairwise compatibility scores for all possible pairs in an outfit to learn the compatibility. 
On the contrary, we use self-attention to learn high-order compatibility relationships between outfit items.
We report the performance of \cite{Cucurull_2019_CVPR} from the paper \cite{coherent2019}
as the authors report performance on Maryland Polyvore \cite{hanACMMM2017} but not on Polyvore Outfits dataset \cite{typeaware2018}. 
\cite{Cucurull_2019_CVPR} requires information from a large number of neighbors in a catalog for best performance, which is impractical for our setting because we do not have prior knowledge about connections between each new item to the existing items, as mentioned in \cite{compretrieval2020}. So, we use k=0 (no neighbors) for a fair comparison.
We observe that using just image features; we outperform other methods that use both image and text features on the compatibility prediction task. 
Using text features boosts performance further.
The methods \cite{typeaware2018,scenet2019,compretrieval2020} employ a pairwise model where they require careful selection of negatives and data augmentation. Our approach uses the outfit compatibility data provided without using any additional strategies and still outperforms the state of the art methods. From Table \ref{tbl:comp}, we observe that transformers can learn better compatibility relationships than other methods \cite{Cucurull_2019_CVPR,hanACMMM2017} that learn compatibility at an outfit-level.

\subsection{FITB and Complementary Item Retrieval (CIR)}
\label{sec:res_retr}

FITB and complementary item retrieval tasks deal with completing an outfit. While for FITB, the task is to select the best item among a fixed set of choices that goes well with an outfit, for retrieval, the task is to choose the best item from the entire database. For the FITB task we use accuracy and for retrieval we use recall@top-k (abbreviated as R@k) as the metric.



Lobert et al.~\cite{lorbert2021} propose to use pre-trained ImageNet embeddings and category for retrieval using self-attention. For evaluation, we adopt their strategy using our own implementation using a transformer \footnote{~Their code is not publicly available and they did not report numbers on the Polyvore Outfits dataset.} and observe that their FITB accuracy on the Polyvore Outfits dataset is 41.61\%. We investigate several strategies such as pre-training on the compatibility prediction task, curriculum learning, a different loss formulation and observe a significant improvement in FITB performance. Our method yields a FITB performance of 58.92\% when using images and category and 67.10 \% using images and text.


\begin{table}[b]
\centering
\caption{Effect of end-to-end training and using different modalities for the CP task.}
\begin{tabular}{@{}lc@{}}
\toprule
\multicolumn{1}{c}{Training strategy} & \multicolumn{1}{l}{CP-AUC} \\ \midrule
ResNet-18 (pre-trained ImageNet) & 0.82 \\
ResNet-18 (end-to-end) & 0.91 \\
ResNet-18 (end-to-end) + Category & 0.92 \\
ResNet-18 (end-to-end) + Text & \textbf{0.93} \\ \bottomrule
\end{tabular}
\label{tbl:comp_mod}
\end{table}
\begin{table*}[t]
\caption{Comparison of different weight initialization schemes and different components of the set-wise outfit ranking loss for the FITB task (using accuracy)}
\centering
\begin{tabular}{@{}lcccc@{}}
\toprule
\multirow{2}{*}{Dataset} & \multicolumn{2}{l}{(a) Pre-training on CP task} & \multicolumn{2}{c}{(b) Ranking Loss Components} \\ \cmidrule(l){2-3} \cmidrule(l){4-5} 
 & without & with & $L_{All}$ & $L_{All} + L_{Hard}$ \\ \midrule
\multicolumn{1}{l}{Polyvore Outfits-D} & 49.15 & \textbf{59.48} & 55.34 & \textbf{59.48} \\
\multicolumn{1}{l}{Polyvore Outfits} & 53.96 & \textbf{67.10} & 64.48 & \textbf{67.10} \\ \bottomrule
\end{tabular}
\label{tbl:abl_retr}
\end{table*}
\begin{table*}[h]
\centering
\caption{Comparison of different strategies for sampling negative samples either from the same high-level or fine-grained category as the positive exemplar for the CIR task.}
\begin{tabular}{@{}lcccccccc@{}}
\toprule
\multirow{2}{*}{Negative sampling} & \multicolumn{4}{c}{Polyvore Outfits-D} & \multicolumn{4}{c}{Polyvore Outfits} \\ \cmidrule(l){2-5} \cmidrule(l){6-9} 
 & \multicolumn{1}{l}{FITB} & \multicolumn{1}{l}{R@10} & \multicolumn{1}{l}{R@30} & \multicolumn{1}{l}{R@50} & \multicolumn{1}{l}{FITB} & \multicolumn{1}{l}{R@10} & \multicolumn{1}{l}{R@30} & \multicolumn{1}{l}{R@50} \\ \midrule
High-level category & 55.54 & 5.14 & 10.21 & 14.15 & 63.33 & 7.26 & 13.60 & 17.78 \\
Fine-grained category & \textbf{59.48} & \textbf{6.03} & \textbf{12.20} & \textbf{16.51} & \textbf{67.10} & \textbf{9.29} & \textbf{16.94} & \textbf{21.82} \\ \bottomrule
\end{tabular}
\label{tbl:abl_negsamp}
\end{table*}
For retrieval, we use the same testing setup as CSA-Net  \cite{compretrieval2020}, and compare the performance of our method with the state of the art methods
CSA-Net \cite{compretrieval2020}, Type-aware \cite{typeaware2018} and SCE-Net average \cite{scenet2019}\footnote{~Type-aware and SCE-Net were adapted for retrieval as reported in \cite{compretrieval2020}.
The performance of \cite{ADDE2021} is not directly comparable to ours, as their method needs to train attributes on another dataset (Shopping100k).}
For evaluation, we use the category as our target item description for retrieving complementary items and use recall@top-k metric that measures the rank of the ground-truth item similar to \cite{compretrieval2020}.
From Table \ref{tbl:retr}, we observe that we outperform all the methods on the non-disjoint dataset. 
On the disjoint dataset, 
%
our performance on recall@top-10 is better than CSA-Net, but is
slightly worse on recall@top-30 and recall@top-50. 
We conjecture that the reason for the performance drop might be because there are fewer outfits on the disjoint set, and transformers typically require large amounts of training data to generalize well.
Also, the authors in CSA-Net discuss that the rank of the ground truth is not a perfect measure for evaluating the retrieval performance. So, we conduct an user study in Section \ref{sec:user_study}.

\subsection{Ablation studies}
\label{sec:ablations}

\noindent
\textbf{Effect of end-to-end training on CP:}
We compare the effect of end-to-end training and using different input modalities on the CP task in Table \ref{tbl:comp_mod}. 
We experiment with using pre-computed ResNet18 image embeddings generated 
as inputs to our transformer model, and observe that training end-to-end improves performance by 9\%. We hypothesize that training end-to-end allows the image encoder to learn better fashion-specific features. 
This allows the transformer to capture visual relationships between items to learn compatibility better. 
We observe that using category or text information further boost performance by 1\% and 2\%, respectively.




\noindent 
\textbf{Pre-training:}
We compare different weight initialization schemes in Table \ref{tbl:abl_retr}(a): 
1) We train our framework for the retrieval task where the transformer-encoder is trained from scratch and the image encoder is initialized with pre-trained ImageNet weights. 
2) We first pre-train our framework on the compatibility prediction task (CP) and then fine-tune our model on the retrieval task. We see a significant improvement in FITB accuracy with the pre-training. 

\noindent 
\textbf{Set-wise outfit ranking loss components:}
As mentioned earlier in Section \ref{sec:ocloss}, the outfit complementarity loss has two components. $L_\text{All}$ optimizes the distances such that the target item embedding is closer to the positive and well separated from the pool of negative samples while $L_\text{Hard}$ focuses specifically on the hard negatives from the randomly sampled pool. From Table \ref{tbl:abl_retr}(b), we  observe that $L_\text{Hard}$ improves FITB performance by 3-4\% on both datasets.



\begin{table*}[h]
\centering
\caption{Comparison of using different input information during inference for CIR.}
\begin{tabular}{@{}lcccccccc@{}}
\toprule
\multirow{2}{*}{Method} & \multicolumn{2}{c}{Input information used} & \multicolumn{3}{c}{Polyvore Outfits-D} & \multicolumn{3}{c}{Polyvore Outfits} \\ \cmidrule(l){2-3} \cmidrule(l){4-6}\cmidrule(l){7-9}
 & Target Item & Outfit Items & \multicolumn{1}{l}{R@10} & \multicolumn{1}{l}{R@30} & \multicolumn{1}{l}{R@50} & \multicolumn{1}{l}{R@10} & \multicolumn{1}{l}{R@30} & \multicolumn{1}{l}{R@50} \\ \midrule
CSA-Net \cite{compretrieval2020} & Catg. & Image + Catg. & 5.93 & \textbf{12.31} & \textbf{17.85} & 8.27 & 15.67 & 20.91 \\
\midrule
OutfitTransf. (Ours) & Catg. & Image + Catg. & 6.03 & 12.20 & 16.51 & 9.29 & 16.94 & 21.82 \\
OutfitTransf. (Ours) & Catg. & Image + Text & \textbf{6.53} & 12.12 & 16.64 & \textbf{9.58} & \textbf{17.96} & \textbf{21.98} \\ \bottomrule
\end{tabular}
\label{tbl:abl_retrmod}
\end{table*}

\noindent
\textbf{Negative sampling strategies:}
In Section \ref{sec:ocloss}, we proposed a curriculum learning strategy where we first sample negatives from the same high-level category as the positive and subsequently sample harder and more informative negatives from the same fine-grained category. This strategy leads to stable training and  improves both FITB and complementary item retrieval performance substantially as shown in Table \ref{tbl:abl_negsamp}. 
Directly training using the hardest negatives from the beginning leads to poor performance.


\noindent
\textbf{Comparing different modalities used for retrieval:}
The OutfitTransformer for retrieval is trained using image and text information. 
Since the complimentary outfit retrieval task in \cite{compretrieval2020} is designed for retrieving items given a target category, we feed the category information to our text encoder and use that for our target item query. We experiment with different modalities during inference, such as using an image with either category information or text description for the items in the partial outfit. From Table \ref{tbl:abl_retrmod}, we observe that when using category information, our method outperforms CSA-Net on the non-disjoint dataset, and using text boosts the retrieval performance further. 

\subsection{Qualitative results}
In this section, we demonstrate qualitatively that our framework retrieves compatible items using target category or text-based descriptions. For this, we use the query outfits from the FITB test split provided by the Polyvore Outfits dataset \cite{typeaware2018}. 

\begin{figure*}[h]
    \centering
    \includegraphics[width=0.9\textwidth]{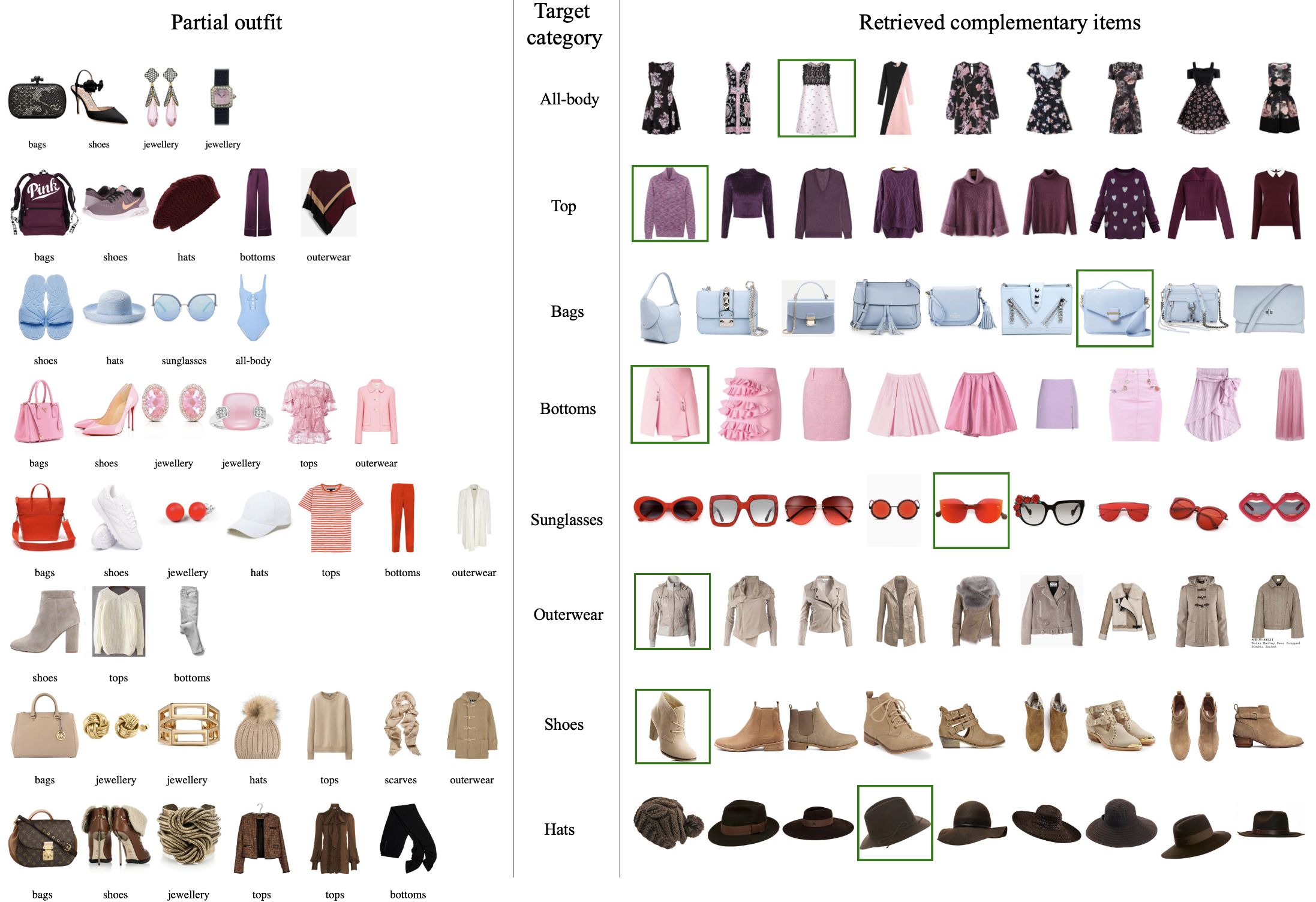}
    \caption{Some example complementary item retrieval results using our framework for different top-level categories. Given a partial outfit and a target category, our method retrieves a list of compatible items that match the global style of the outfit. The ground truth item is indicated by the green bounding box. }
    \label{fig:CatgQuery}
\end{figure*}

For the complementary item retrieval task based on a target category, we use the dataset proposed by Lin \etal \cite{compretrieval2020}. 
In Fig. \ref{fig:CatgQuery}, we show examples where our method can successfully retrieve the ground truth as one of the top retrieved items. Subsequently, in Fig. \ref{fig:fail_CatgQuery} we show examples where the the ground truth item is not found in the top retrieved results. We observe that this is mainly because there can be multiple items in the dataset that could be a good match for the partial outfit either because it is either very similar in style, color, etc. to the ground truth item as shown in Fig. \ref{fig:fail_CatgQuery}(a) or it might be dissimilar to the ground truth but can still be compatible with the overall outfit as shown in Fig. \ref{fig:fail_CatgQuery}(b). Therefore, the rank of the ground truth item is not a perfect indicator of the practical utility of the system. This is a limitation of the evaluation metric due to the lack of annotated ground truth as mentioned in \cite{compretrieval2020}. Towards this end, we conduct a user study as described in Section \ref{sec:user_study} to further validate the quality of the retrieval results and for a better understanding of the usefulness of our framework. 

\begin{figure*}[h]
   \centering
   \begin{minipage}[c]{0.43\textwidth}
   \includegraphics[width=\textwidth]{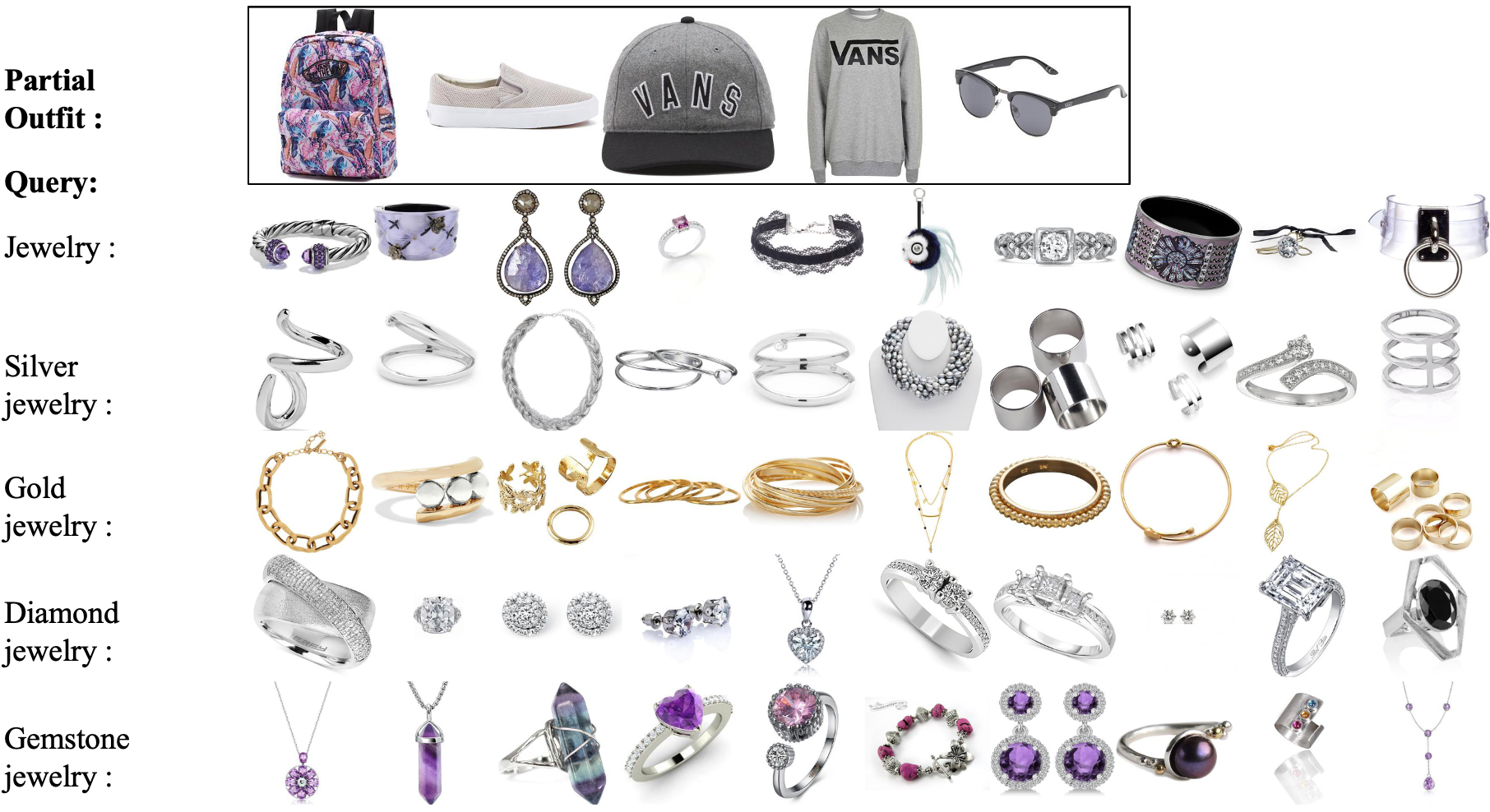}
   \includegraphics[width=\textwidth]{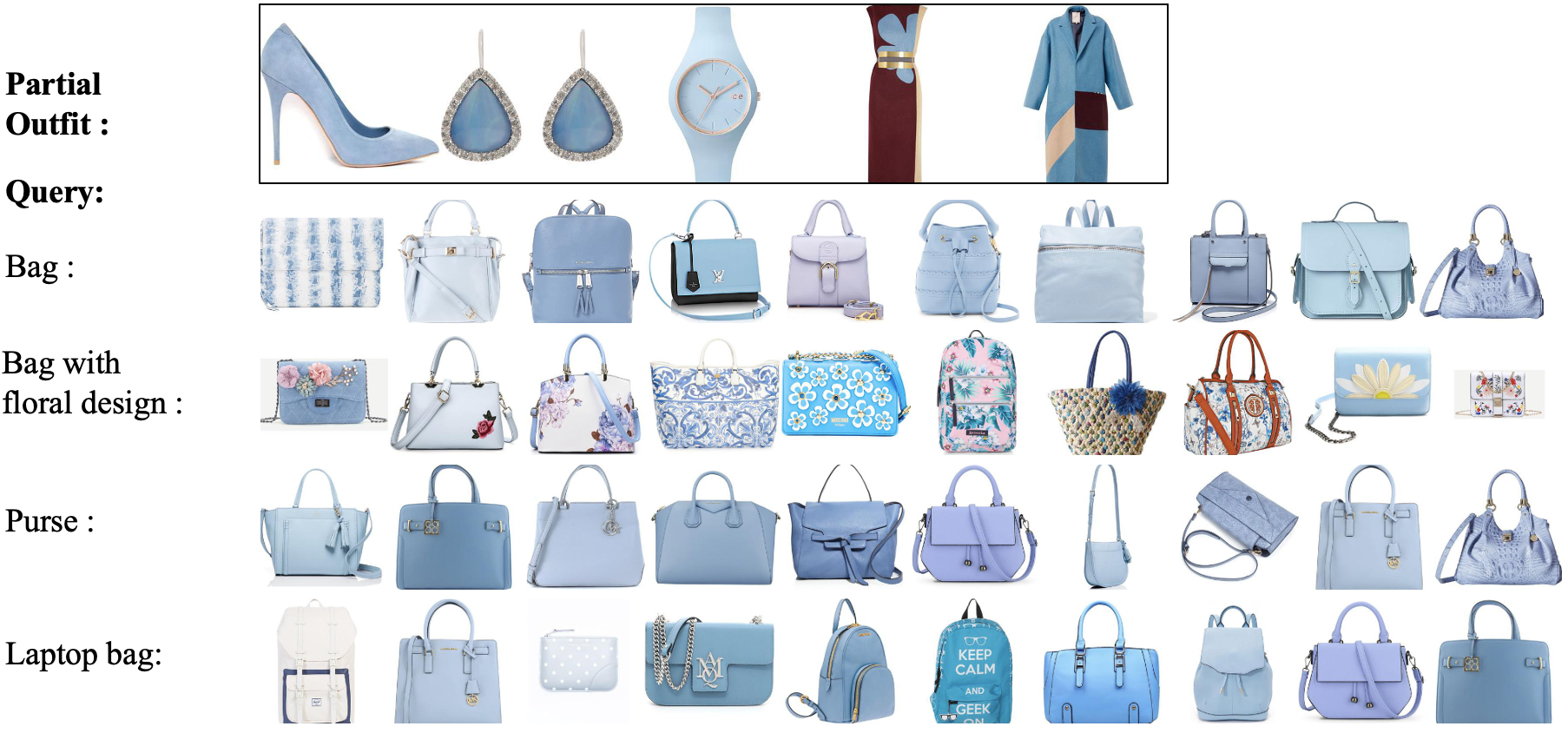}
   \caption*{(a) Text-based queries}
   \end{minipage}
   \hfill \vline \hfill 
   \begin{minipage}[c]{0.47\textwidth}
   \includegraphics[width=\textwidth]{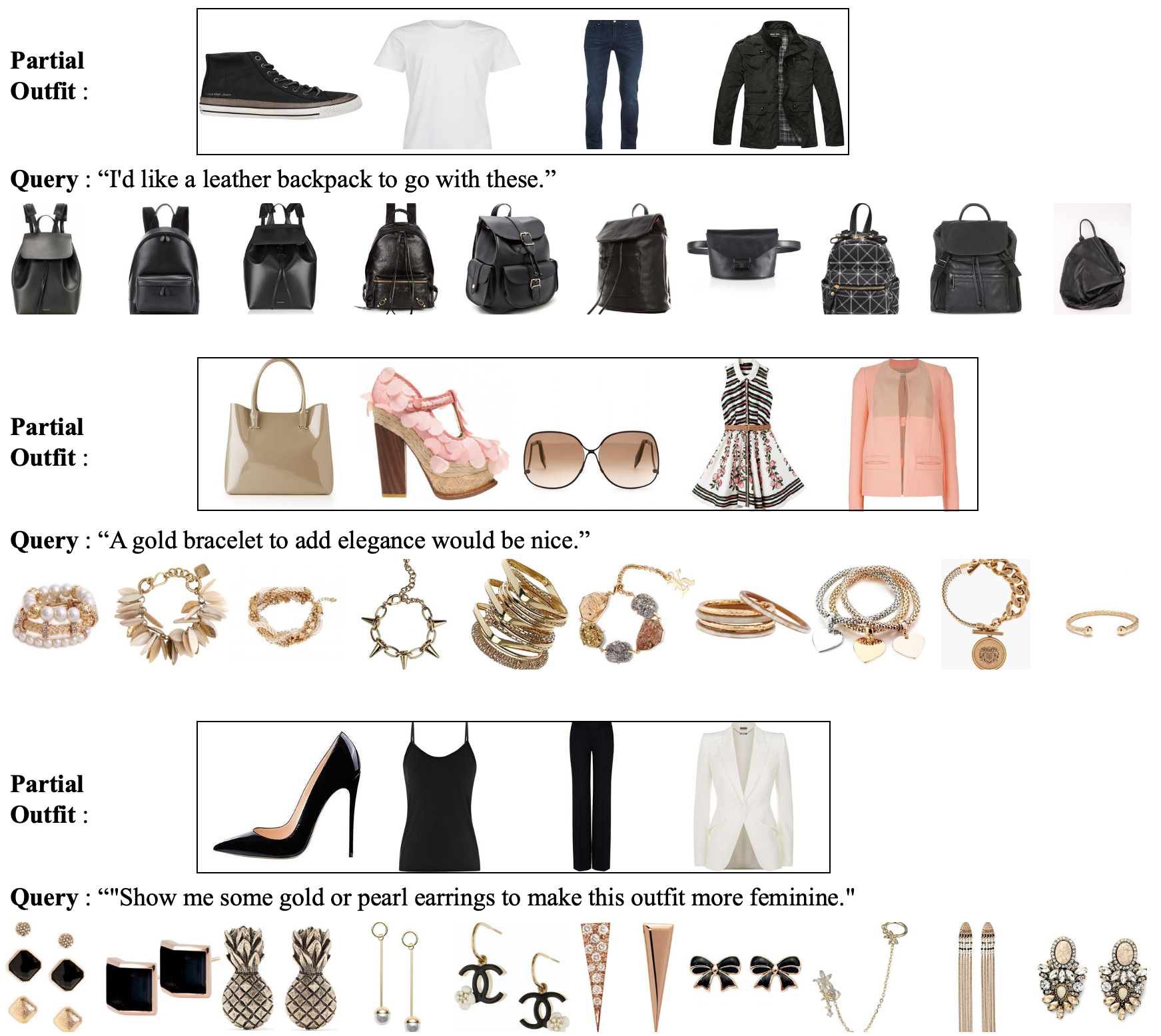}
   \caption*{(b) Conversational queries}
   \end{minipage}
   \caption{Some example complementary item retrieval results using our method for different target item descriptions. Given a partial outfit and text-based queries, our approach retrieves a list of compatible items that match both the global style of the outfit and the text description.}
   \label{fig:textretr}
\end{figure*}

\begin{figure*}[h]
    \centering
    \includegraphics[width=\textwidth]{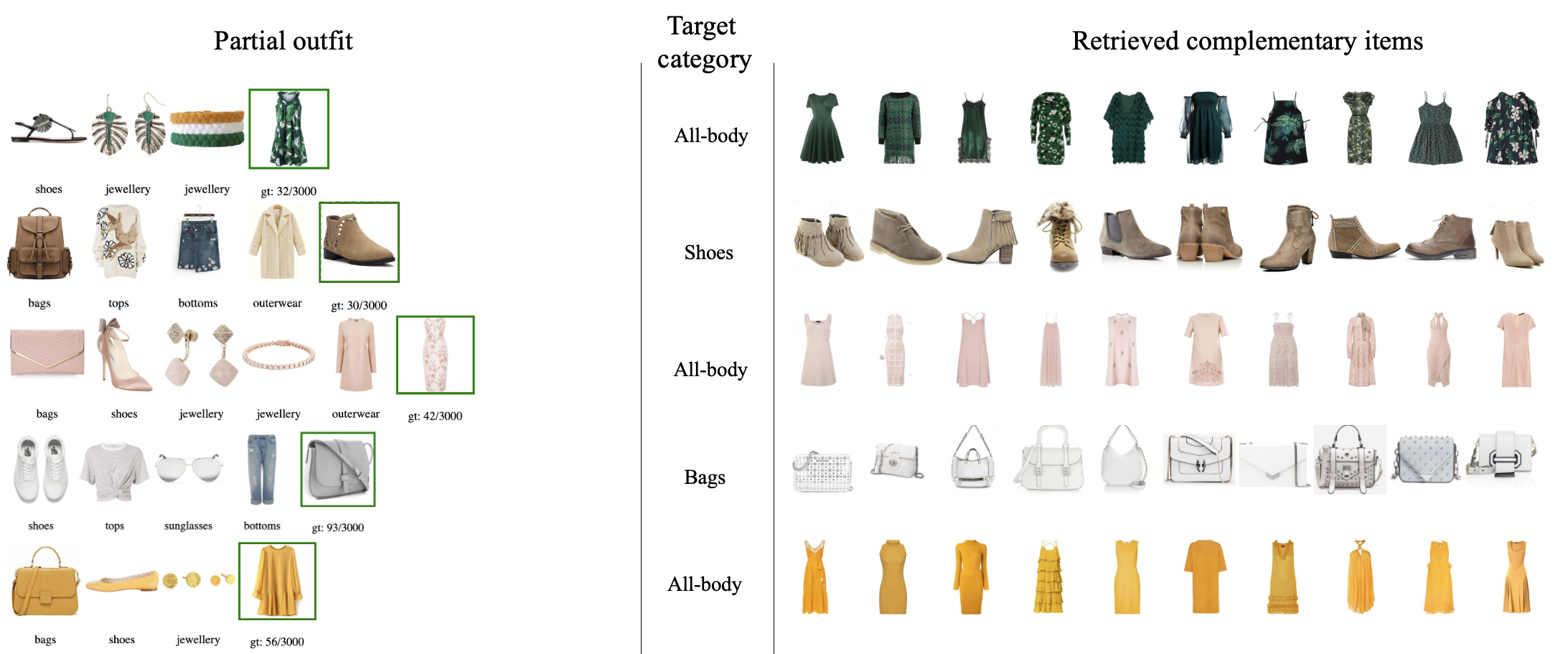}
    \caption*{(a) Retrieved items are very similar in color, texture, style, pattern, etc. to the ground truth item.}
    \vspace{0.25in}
    \includegraphics[width=\textwidth]{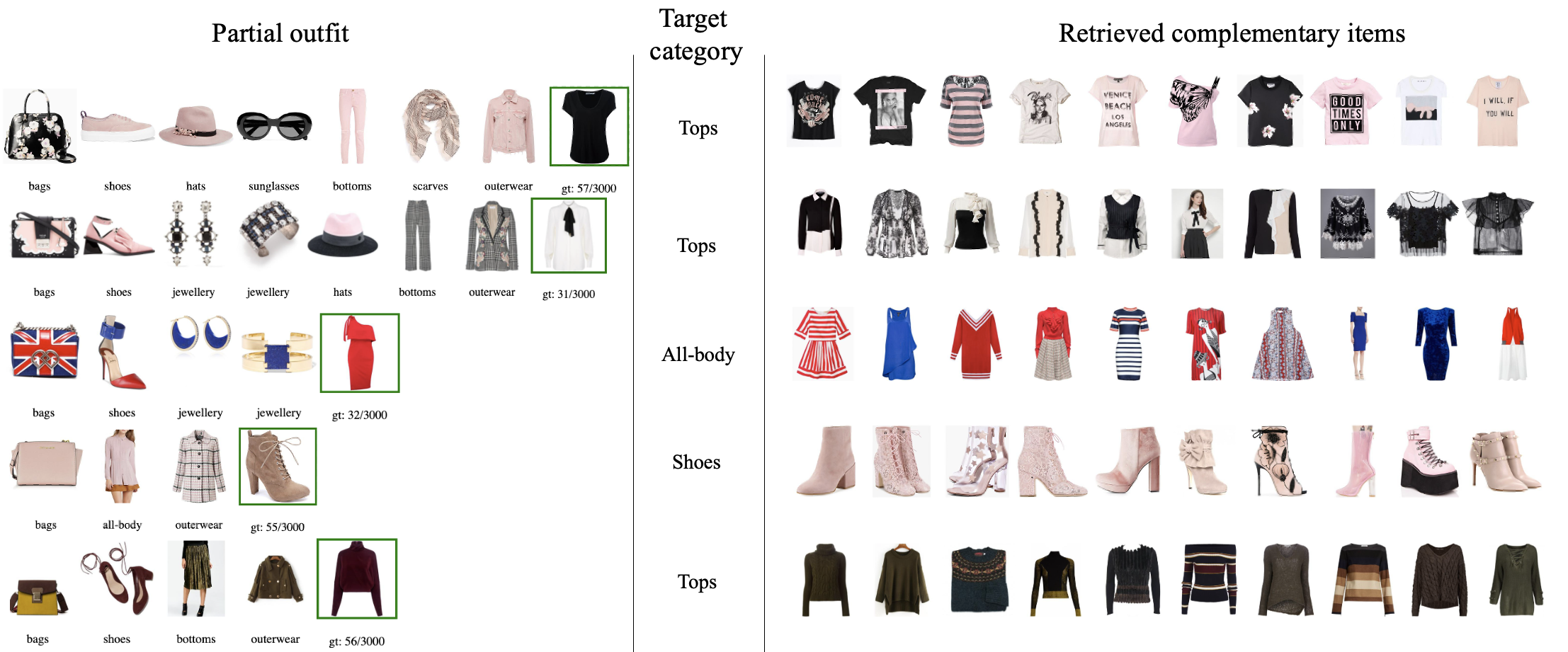}
    \caption*{(b) Retrieved items are different in color, style, etc. from the ground truth item but they are also compatible with the partial outfits.}
    \caption{Some failure cases of complementary item retrieval using our method. The partial outfit and the corresponding ground truth item are shown in the left column, the target category in the middle column and the retrieved items are shown in the right column. The ground truth is highlighted using a green bounding box and the rank of the item is also presented below the box.}
    \label{fig:fail_CatgQuery}
\end{figure*}

We also use our framework for retrieving complementary items based on a target item description provided as free-form text, as shown in Figs. \ref{fig:textretr} and \ref{fig:jewelryretr}. 
Since the dataset does not provide any annotated text-based queries, we show qualitative results that demonstrate that our framework can retrieve items that are {\em both} compatible and matches the target item description in Fig. \ref{fig:textretr}. In Fig. \ref{fig:jewelryretr}, we show that for a given list of text-based queries and different partial outfits, our system can retrieve complementary items for each specific outfit.
\subsection{User Study}
\label{sec:user_study}

The recall@top-k metrics reported in the paper rely on the relative rank of the ground-truth item in the complementary item retrieval task. While high values on these metrics indicate that our model performs competitively or better than state-of-the-art, as \cite{compretrieval2020} point out, the rank of the ground-truth item is not a perfect indicator of retrieval performance since the database can contain many complementary items to the query outfits -- some of which may be judged by human experts to be equally-good or even better stylistic matches, as can be seen in Fig. \ref{fig:fail_CatgQuery}.

We evaluate this hypothesis by running an A/B test using Amazon Mechanical Turk (c.f. Fig. \ref{fig:UserStudy} showing the question template used  for the experiment). If human experts select our retrieved item over the ground truth item about 50\% of the time or higher\footnote{higher than 50\% would indicate that for some partial outfits, our method can find a better match from the database of items belonging to the target category than the original answer provided by the dataset}, that would indicate that even in cases where the ground-truth item is not found in the top-k retrieval results, the retrieved items are a good replacement. 

For quality control of our annotator pool, we only allow annotators with $>$ 95\% acceptance rate in the past tasks and who have completed at least 1000 tasks.  A query outfit from the FITB test split of the Polyvore Outfits dataset is presented, and the annotator is asked to select which of the two items provided better completes the outfit (as shown in Fig. \ref{fig:UserStudy}). We only present questions where the ground-truth item is not found in the top-three retrieved items. One of the presented items is always the ground-truth, and the other — an item retrieved by our model. The total number of questions is 5120, each of which is annotated by 10 subjects. We compute the average number of times the retrieved item was chosen over the ground truth item. We find that 52.5\% of the time annotators select items retrieved by our model over the ground truth. The average rate of 52.5\% denotes that our retrieved results were at least as good as the ground truth items. This provides an additional evidence over and above the offline results that our model is able to retrieve compatible items to the query outfit, and they are as good of a match as the ground truth item. 

\section{Conclusion}
We present a framework to learn outfit-level representations for compatibility prediction and complementary item retrieval.
Experimental results demonstrate that our model outperforms several state-of-the-art approaches on the Polyvore Outfits dataset in three established tasks: outfit compatibility, fill-in-the-blank, and complementary item retrieval. We validate that our retrieved results are competitive with the ground truth via a user study, and demonstrate qualitatively that our framework retrieves compatible items using target category or text-based descriptions. In future work, we plan to extend complementary item retrieval to sets of items rather than one-at-a-time.

\newpage

\begin{figure*}[h]
    \centering
    \begin{minipage}[c]{0.49\textwidth}
   \includegraphics[width=\textwidth]{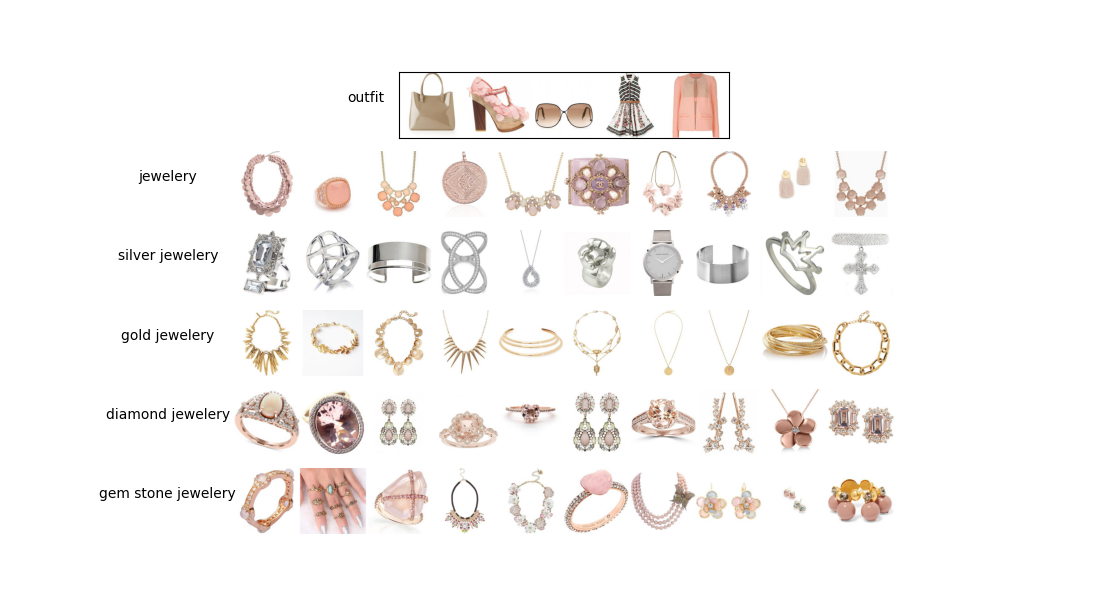}
   \includegraphics[width=\textwidth]{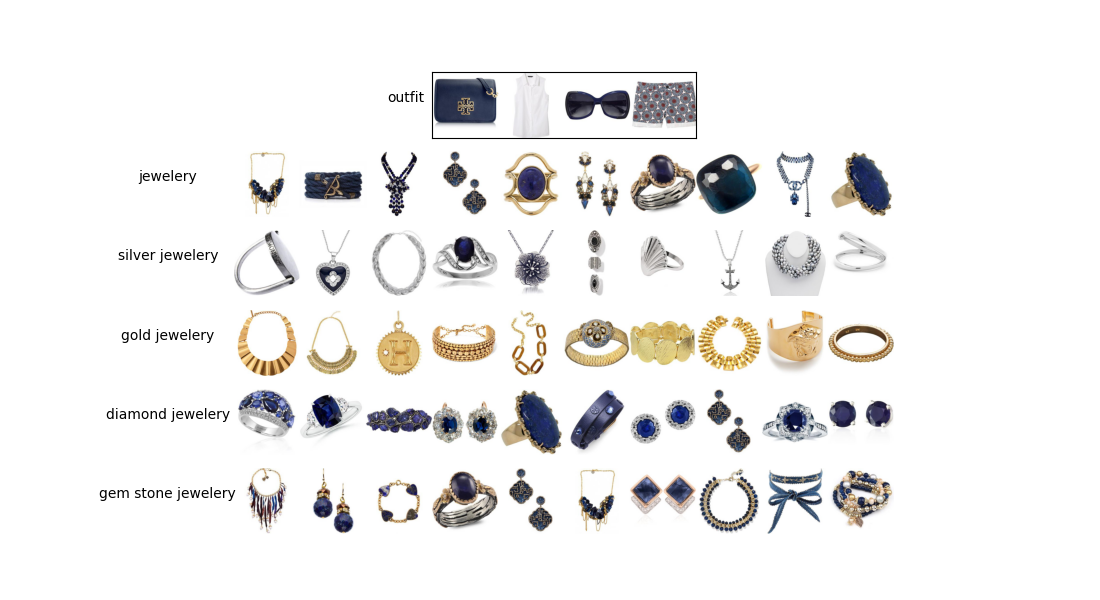}
   \includegraphics[width=\textwidth]{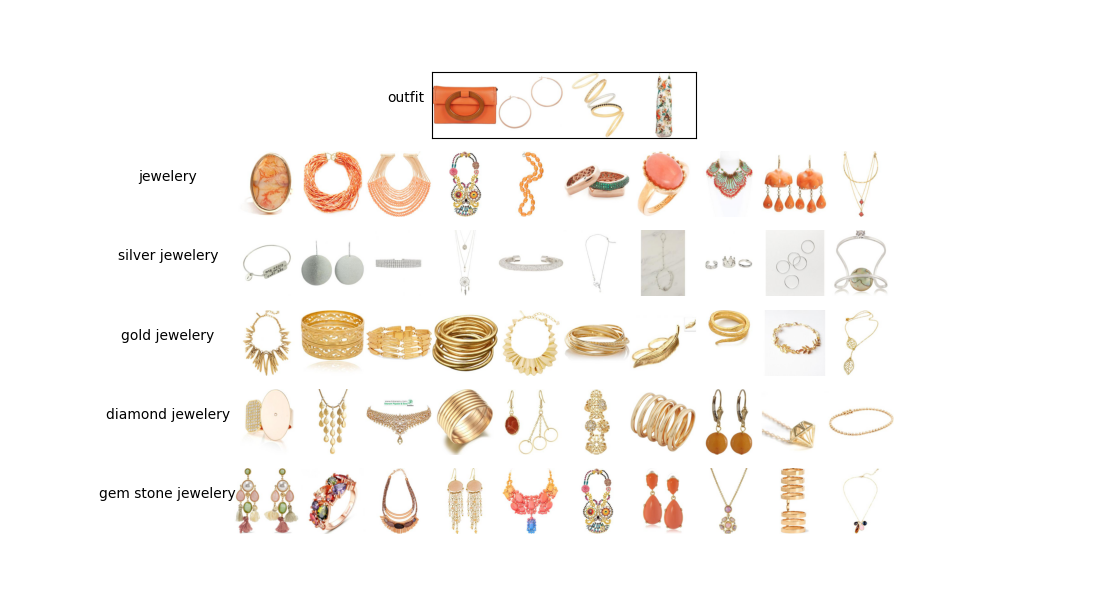}
   \includegraphics[width=\textwidth]{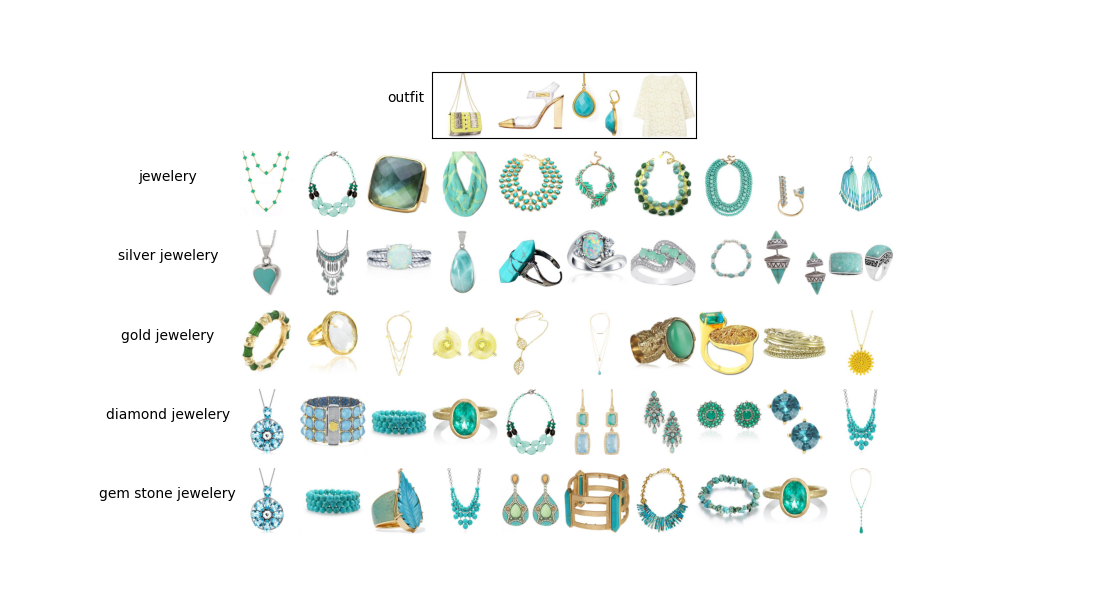}
   \includegraphics[width=\textwidth]{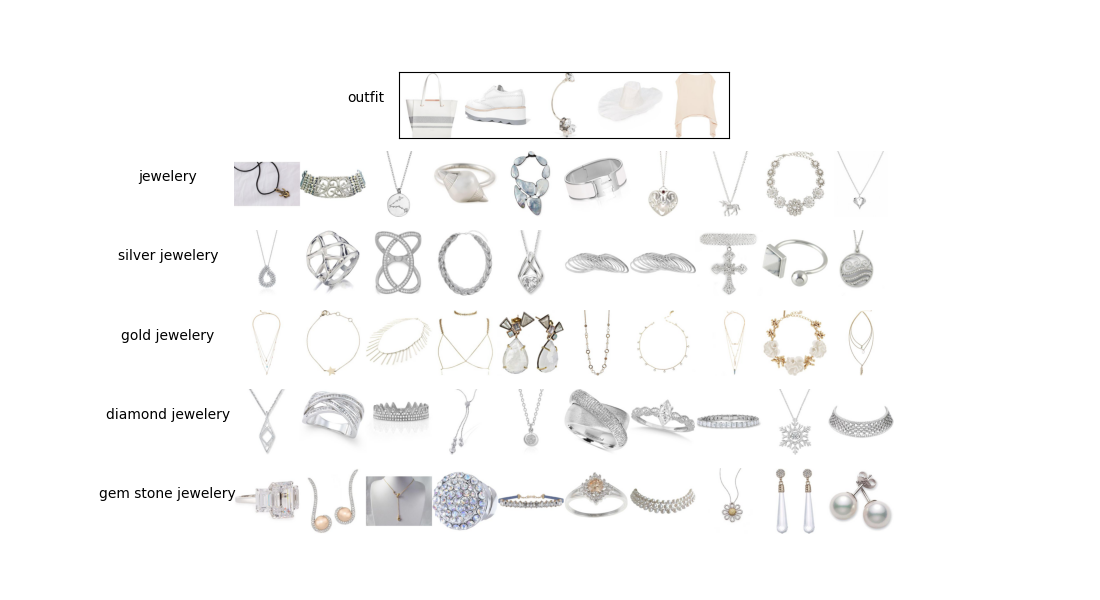}
   \end{minipage}
   \hfill 
   \begin{minipage}[c]{0.49\textwidth}
   \includegraphics[width=\textwidth]{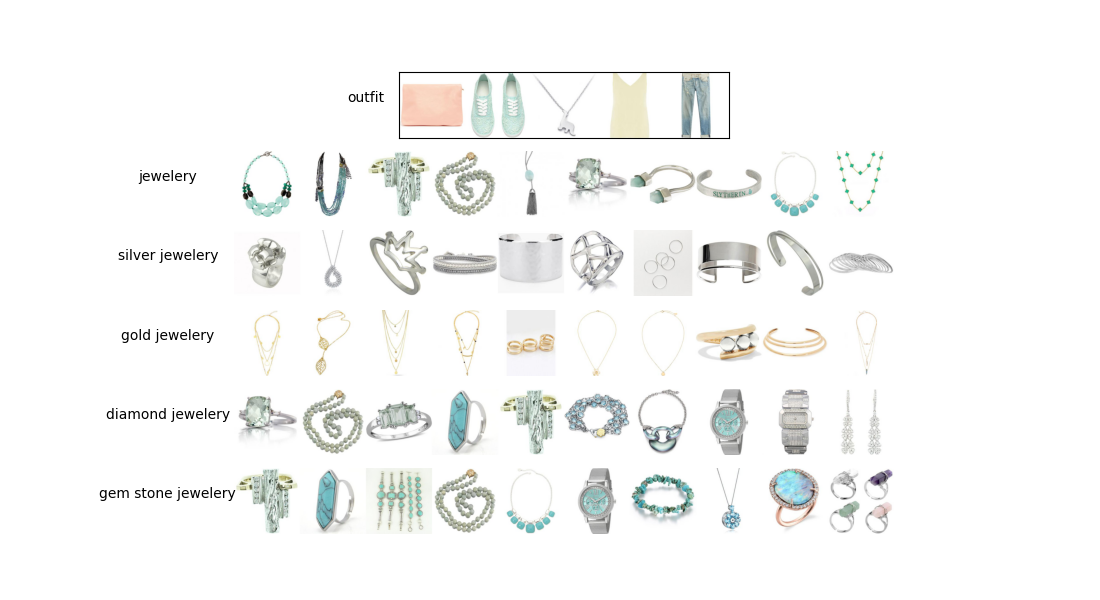}
   \includegraphics[width=\textwidth]{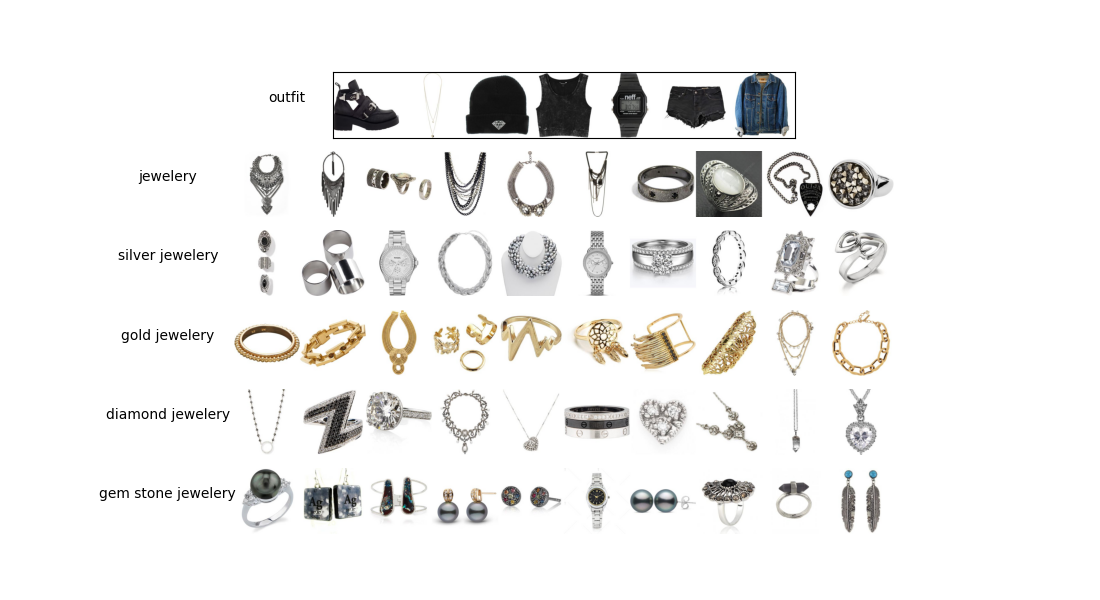}
   \includegraphics[width=\textwidth]{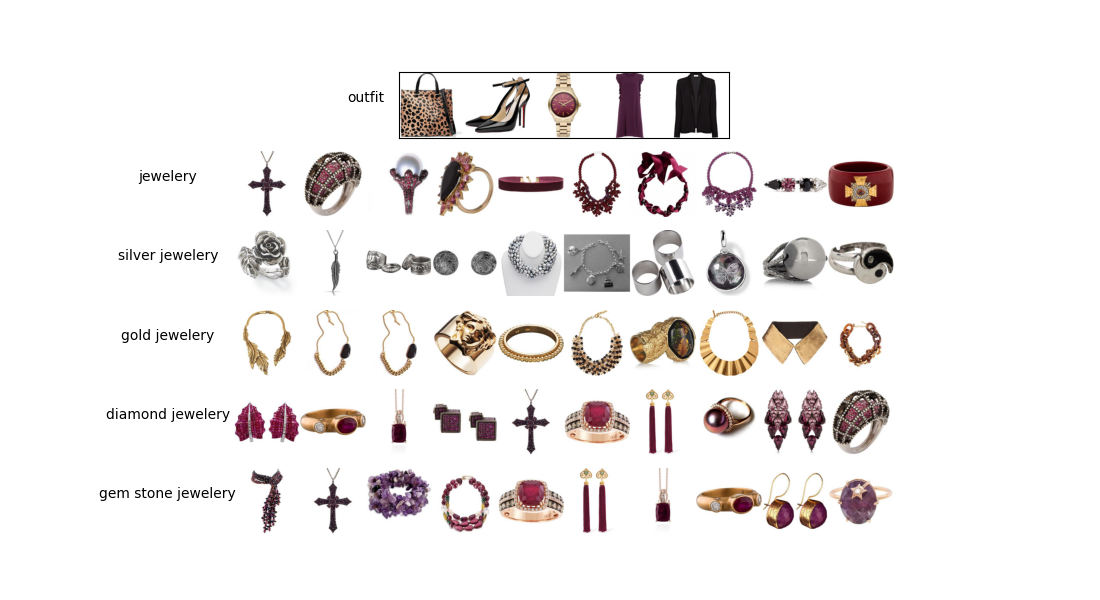}
   \includegraphics[width=\textwidth]{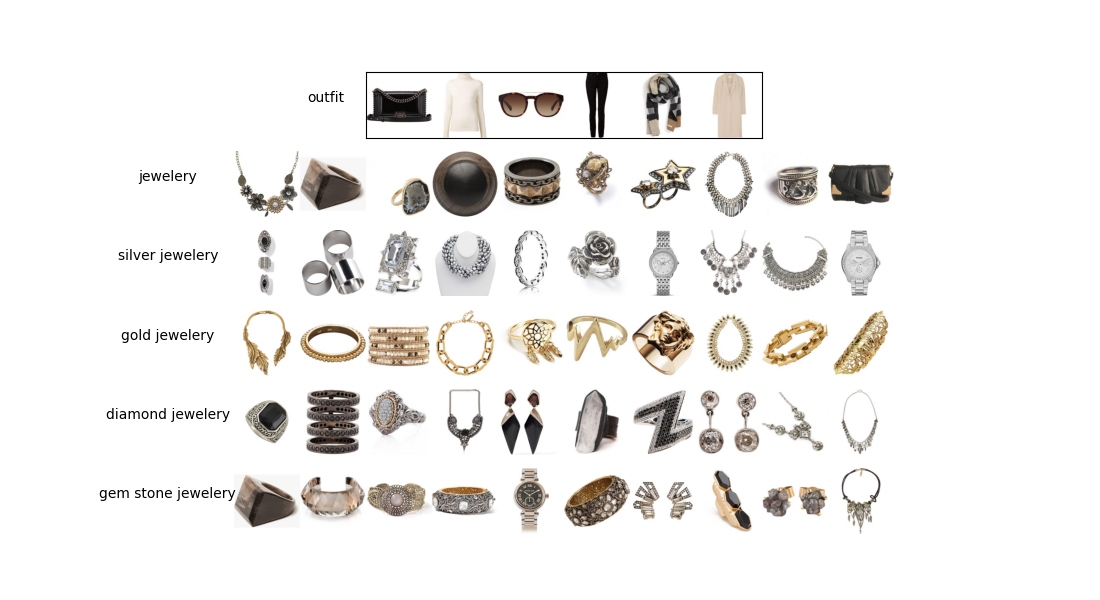}
   \includegraphics[width=\textwidth]{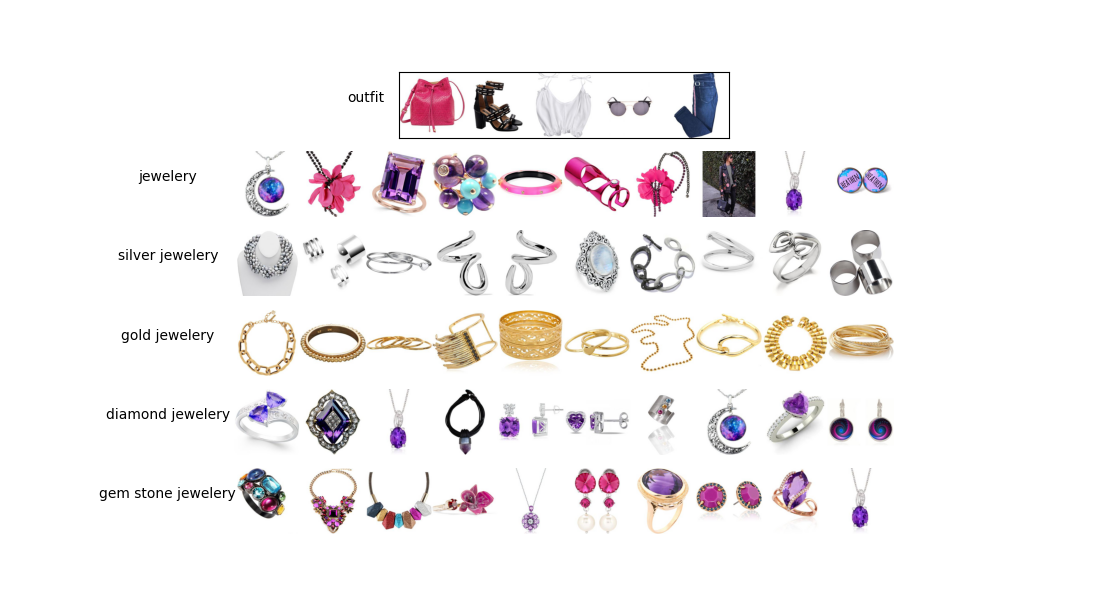}
   \end{minipage}
   \end{figure*}
  \begin{figure*}[h]
    \centering
   \begin{minipage}[c]{0.49\textwidth}
   \includegraphics[width=\textwidth]{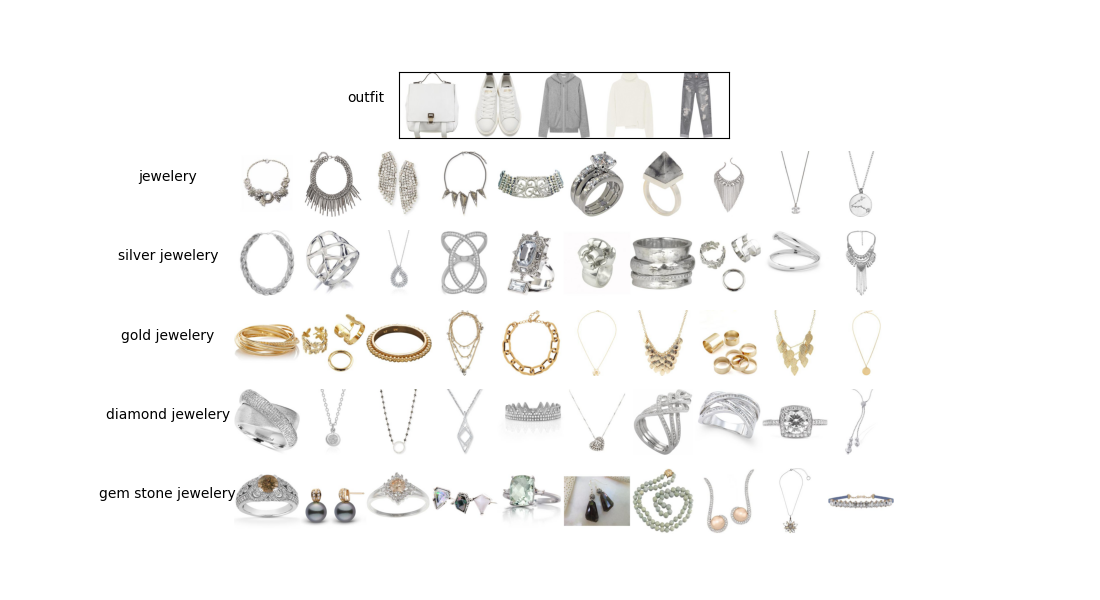}
   \includegraphics[width=\textwidth]{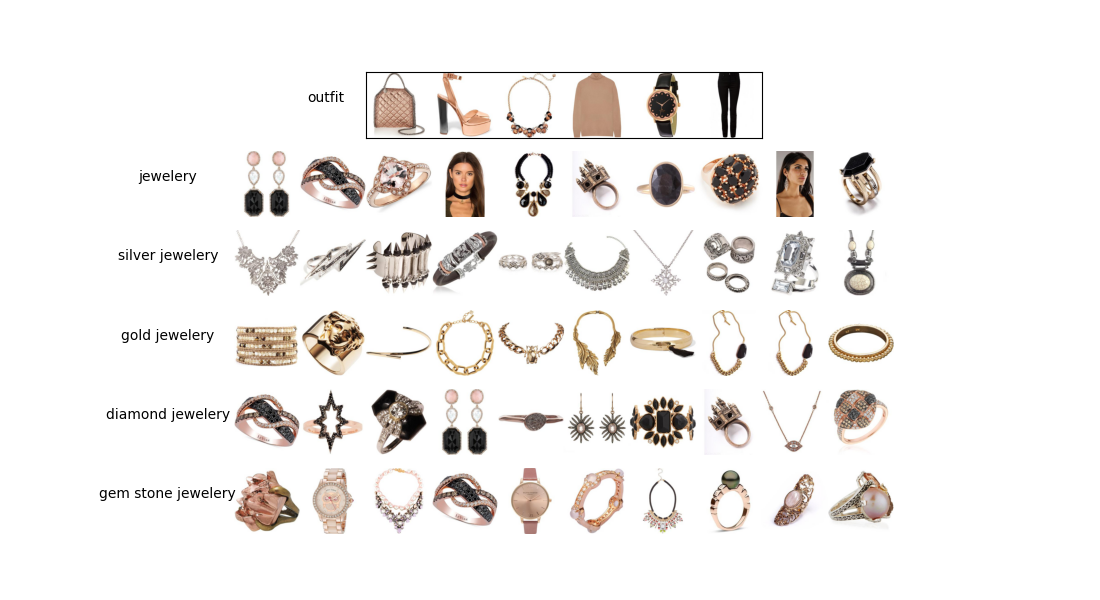}
   \includegraphics[width=\textwidth]{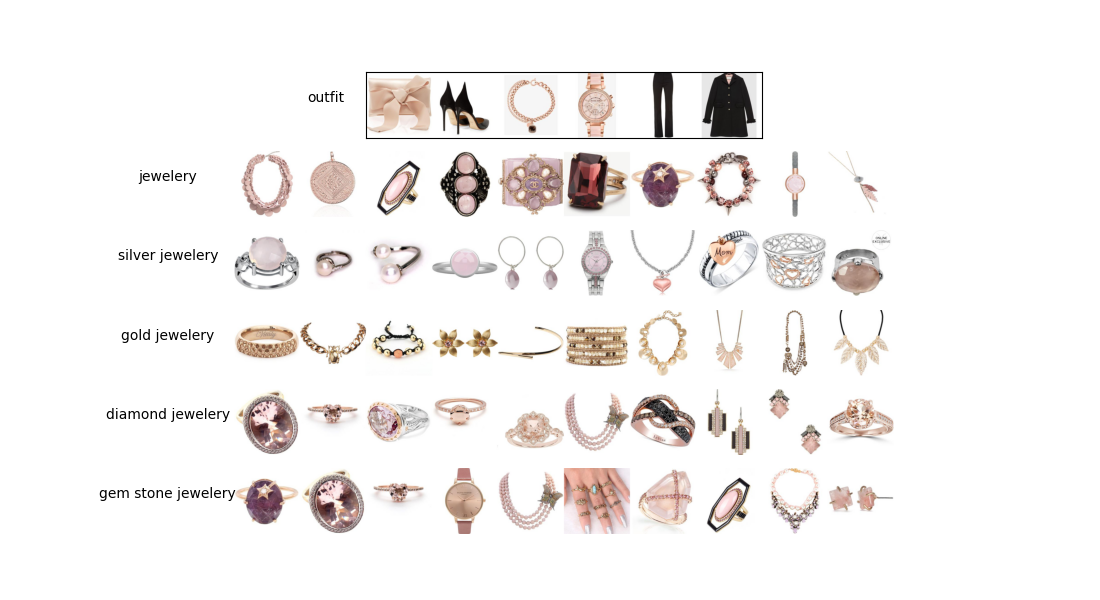}
   \includegraphics[width=\textwidth]{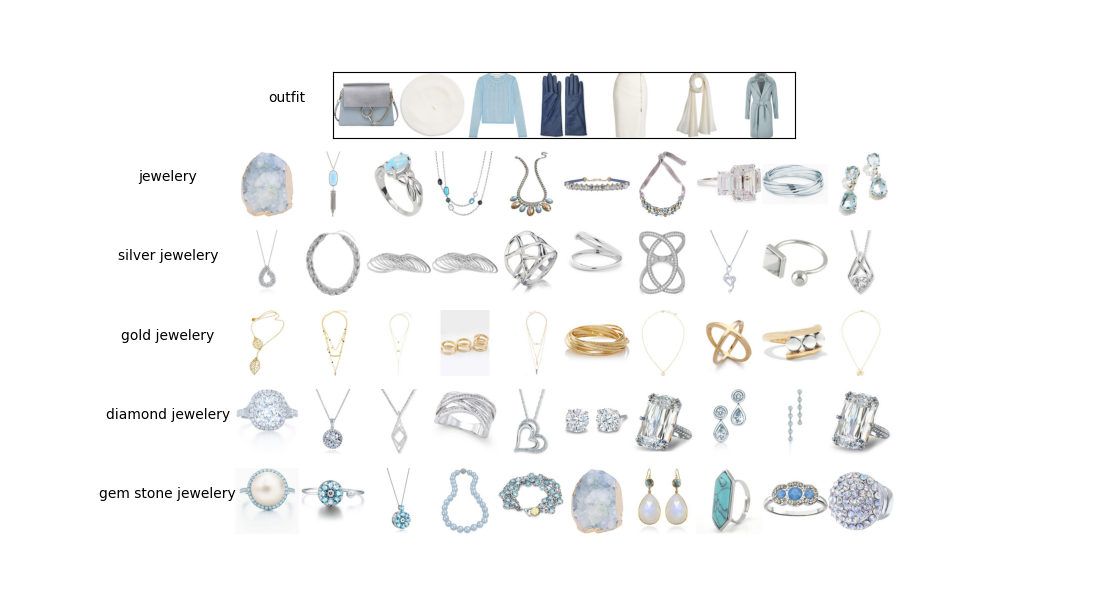}
   \end{minipage}
   \begin{minipage}[c]{0.49\textwidth}
   \includegraphics[width=\textwidth]{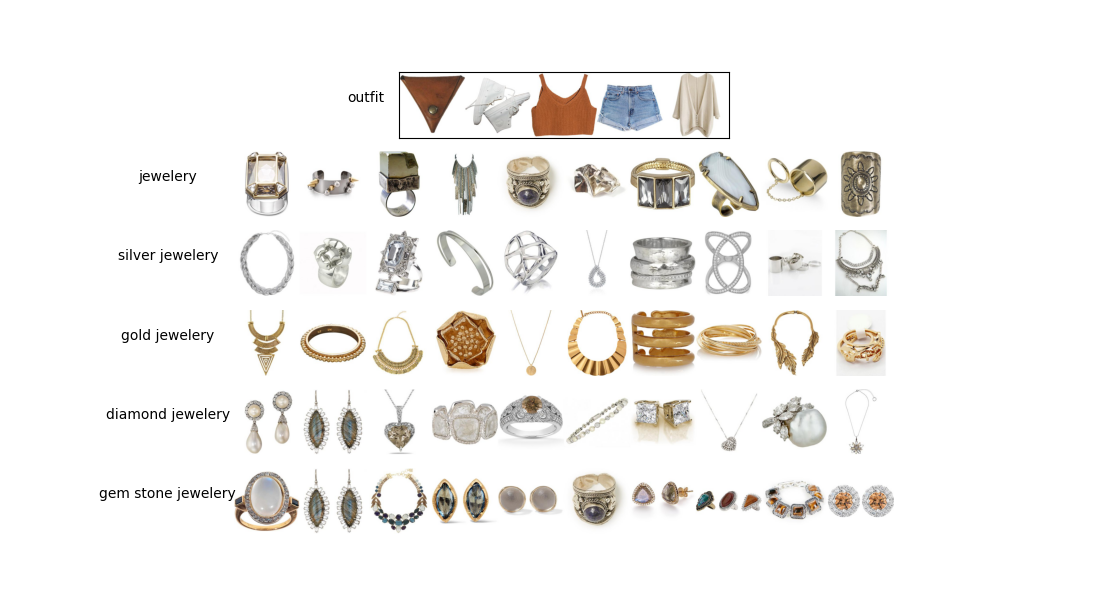}
   \includegraphics[width=\textwidth]{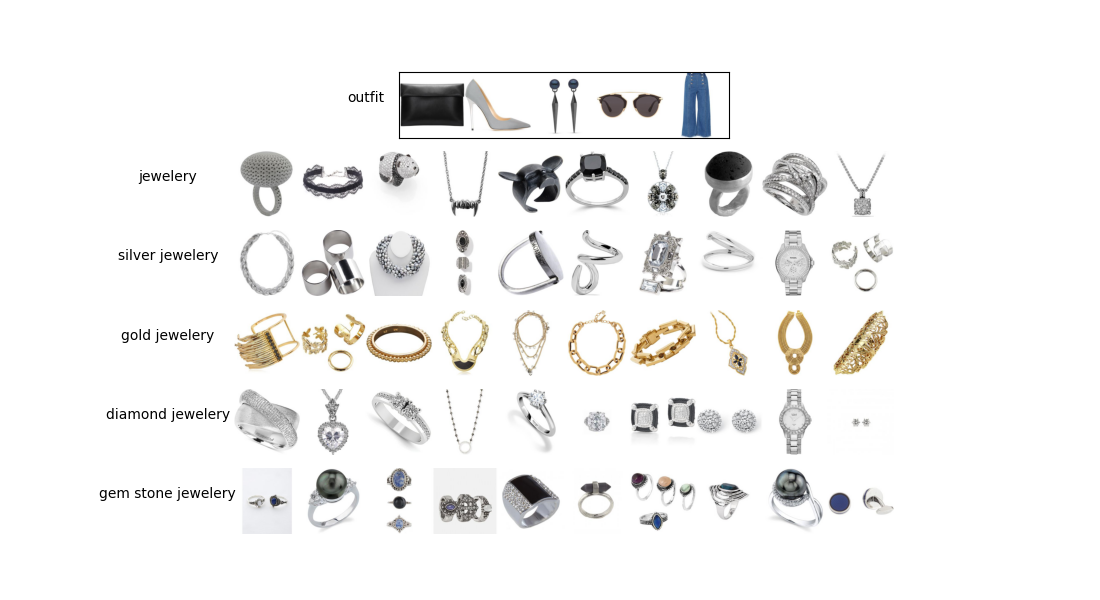}
   \includegraphics[width=\textwidth]{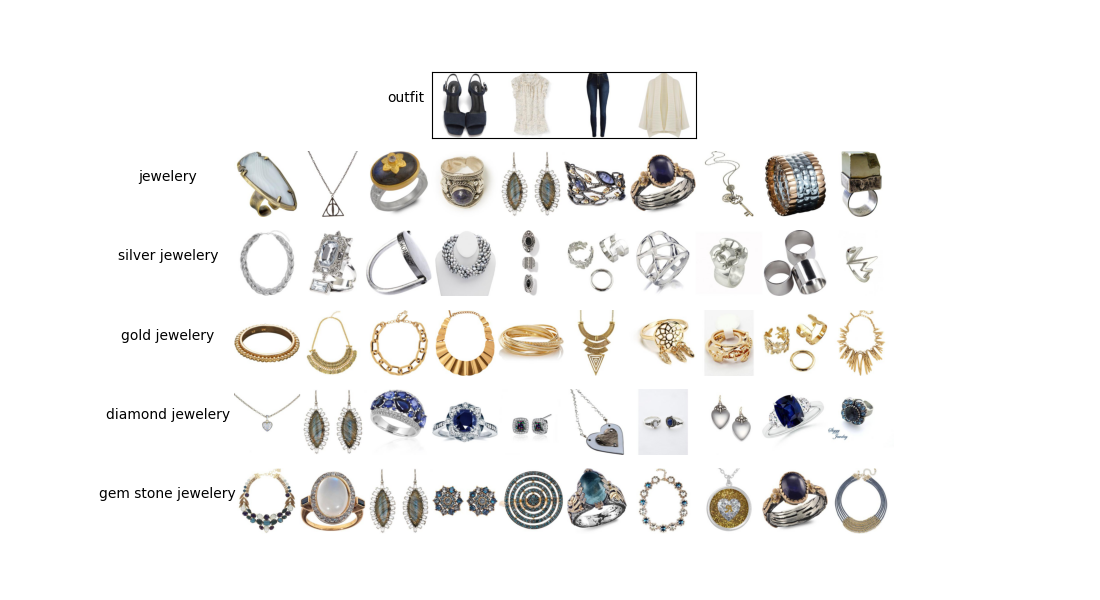}
   \includegraphics[width=\textwidth]{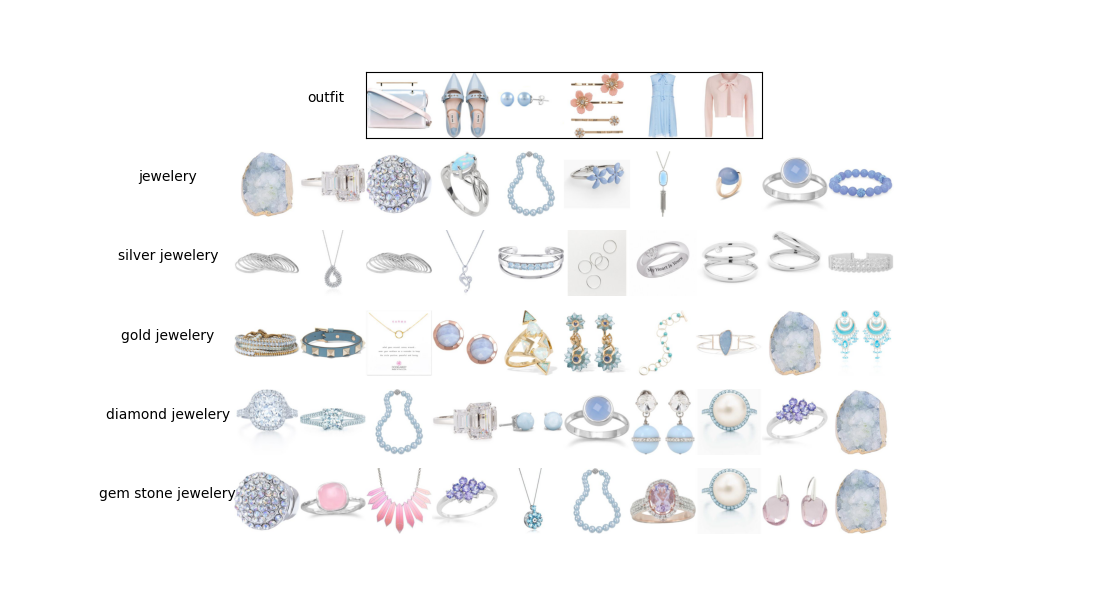}
   \end{minipage}
    \caption{This figure illustrates that our method can retrieve items that are compatible with different partial outfits for the same list of text-based queries. The first query for each partial outfit mentions only the top-level category information whereas the subsequent queries are more descriptive which allows us to further refine search results. }
    \label{fig:jewelryretr}
\end{figure*}

\begin{figure*}[tp]
    \centering
    \begin{minipage}[c]{0.49\textwidth}
   \includegraphics[width=\textwidth]{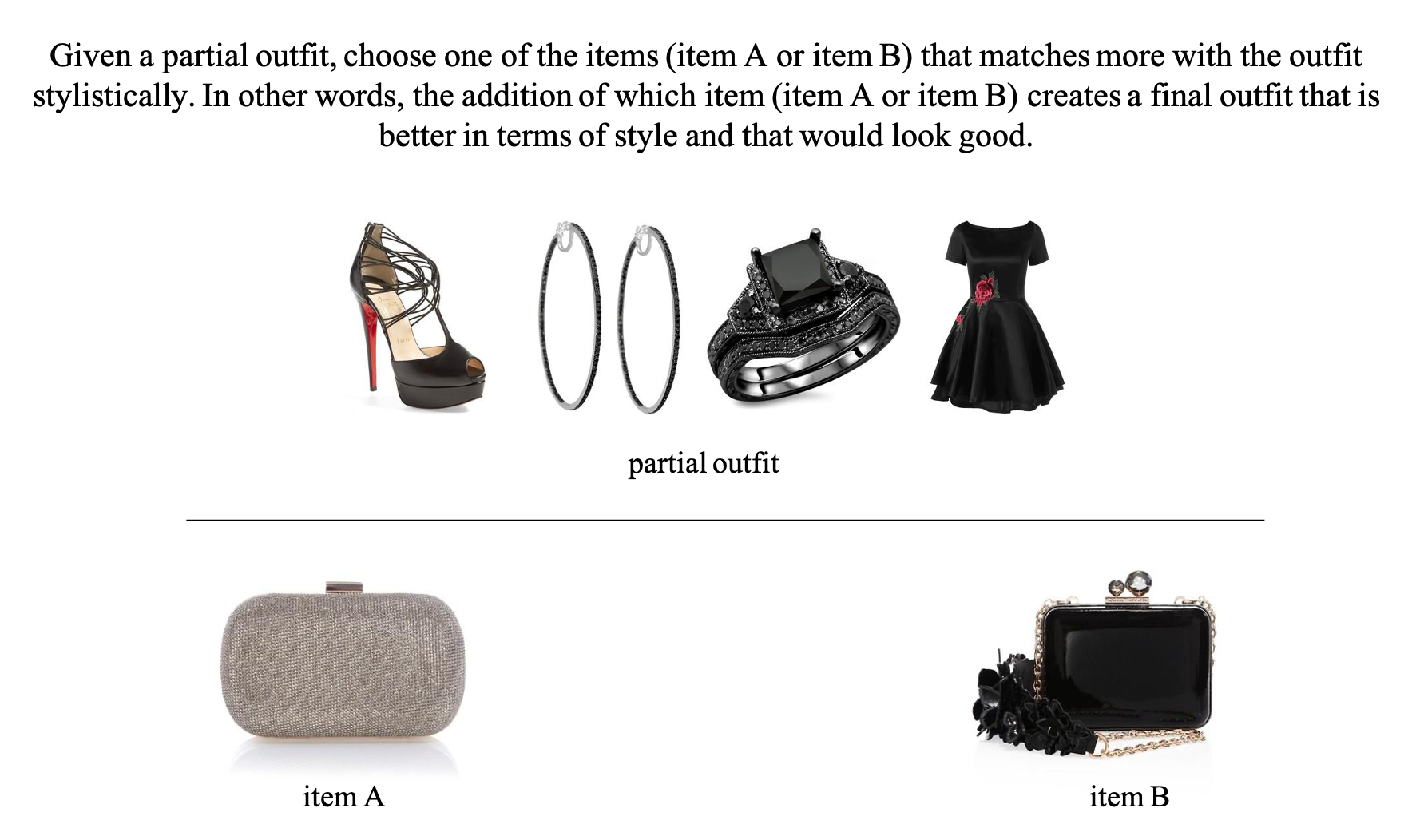}
   \includegraphics[width=\textwidth]{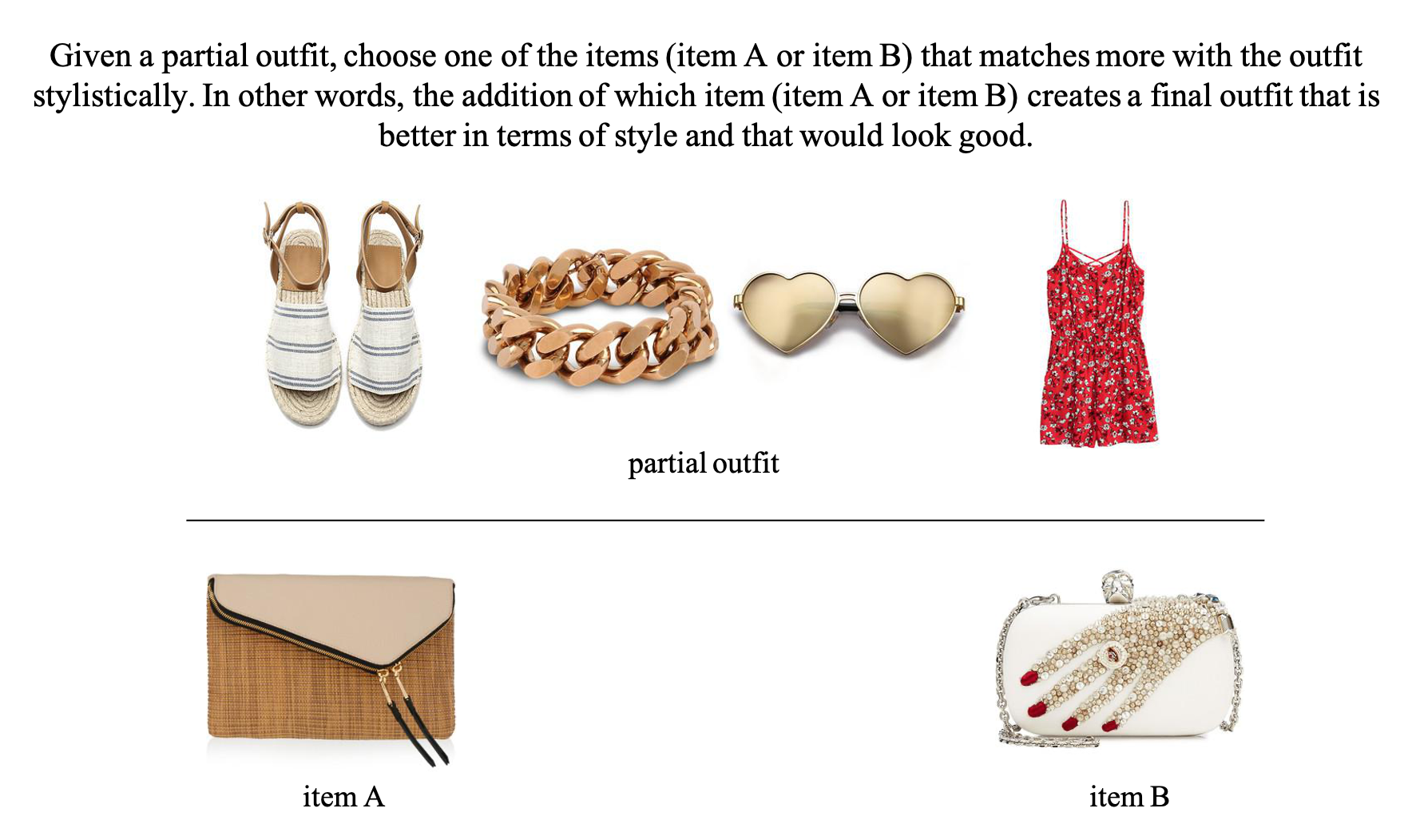}
   \includegraphics[width=\textwidth]{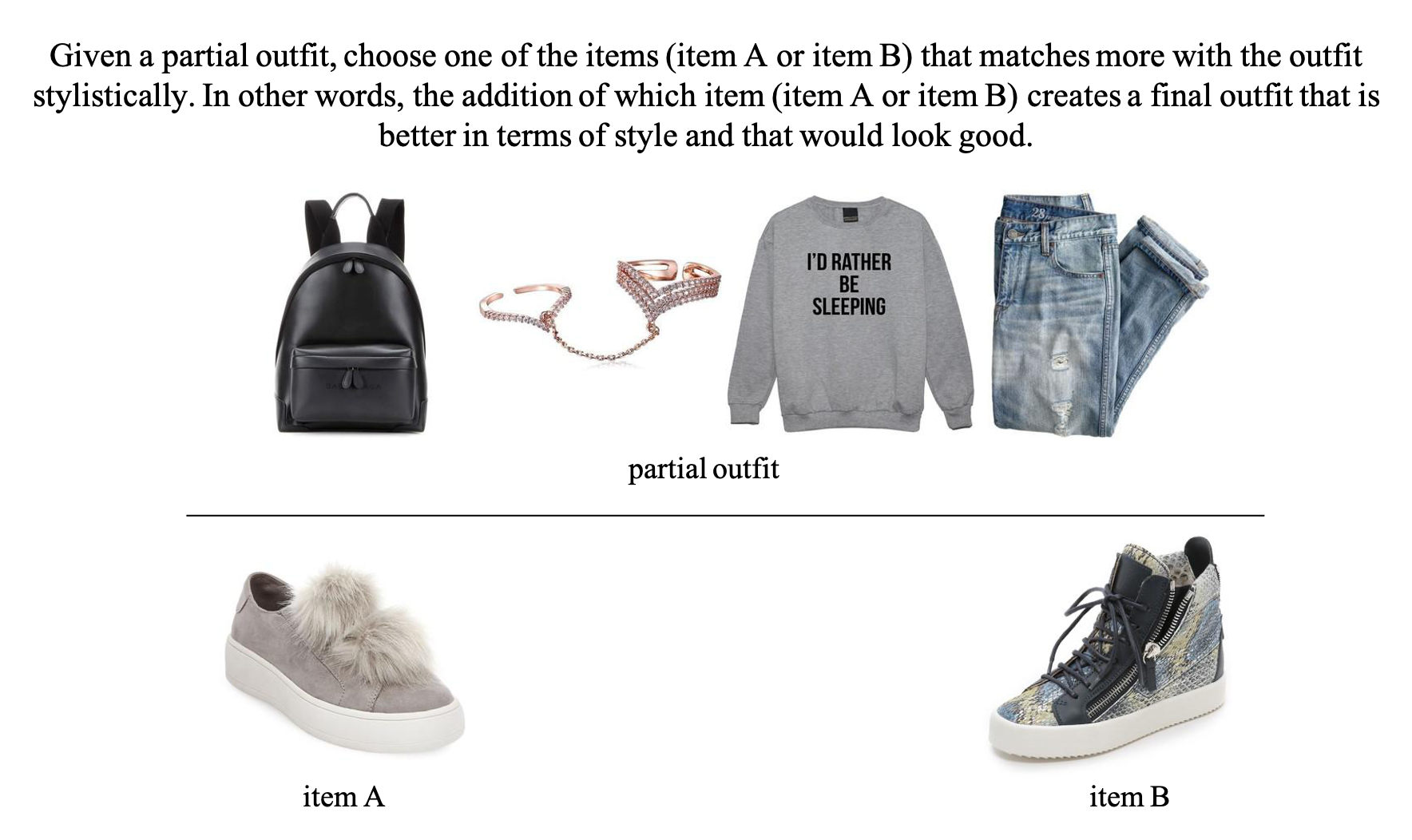}
   \end{minipage}
   \hfill 
   \begin{minipage}[c]{0.49\textwidth}
   \includegraphics[width=\textwidth]{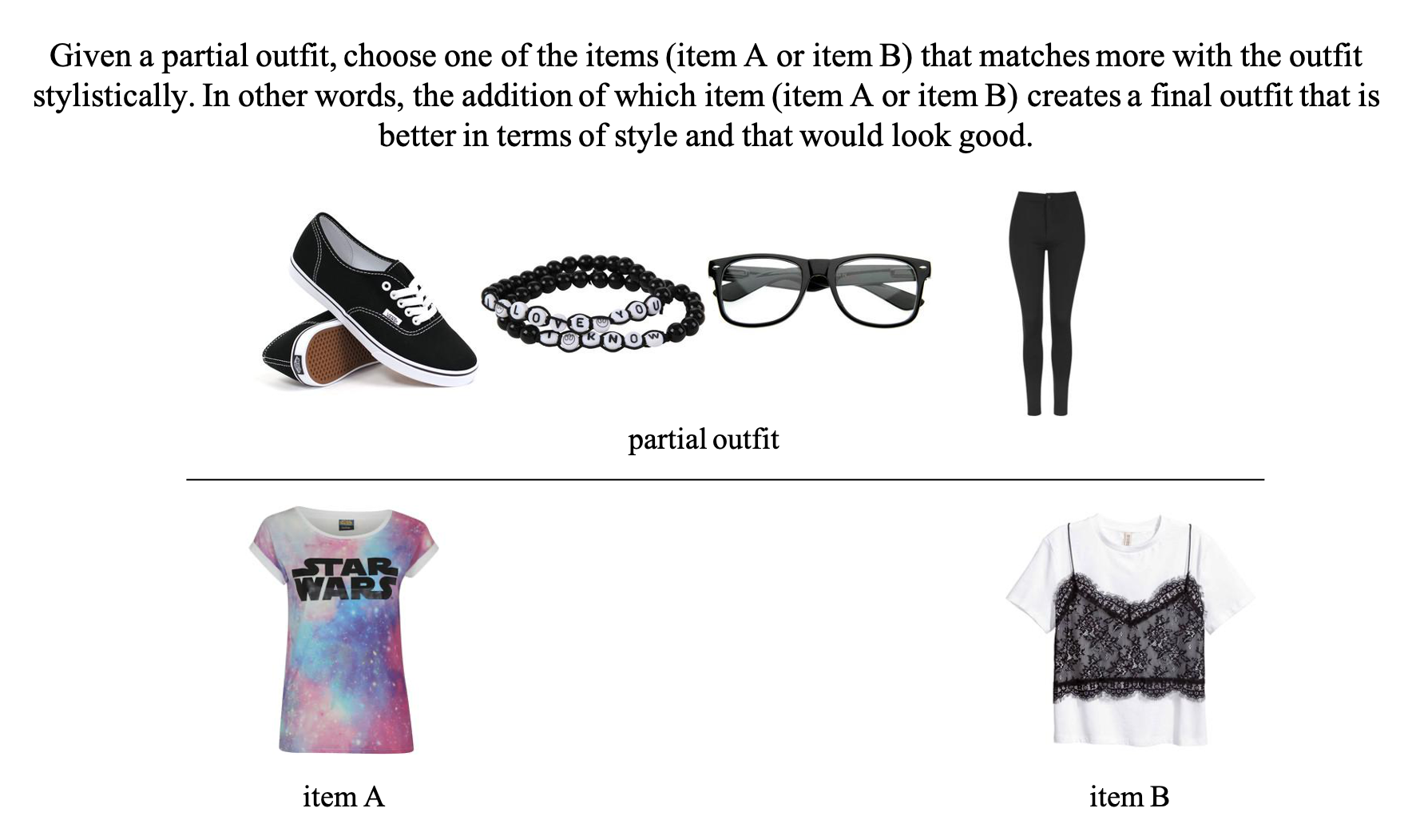}
   \includegraphics[width=\textwidth]{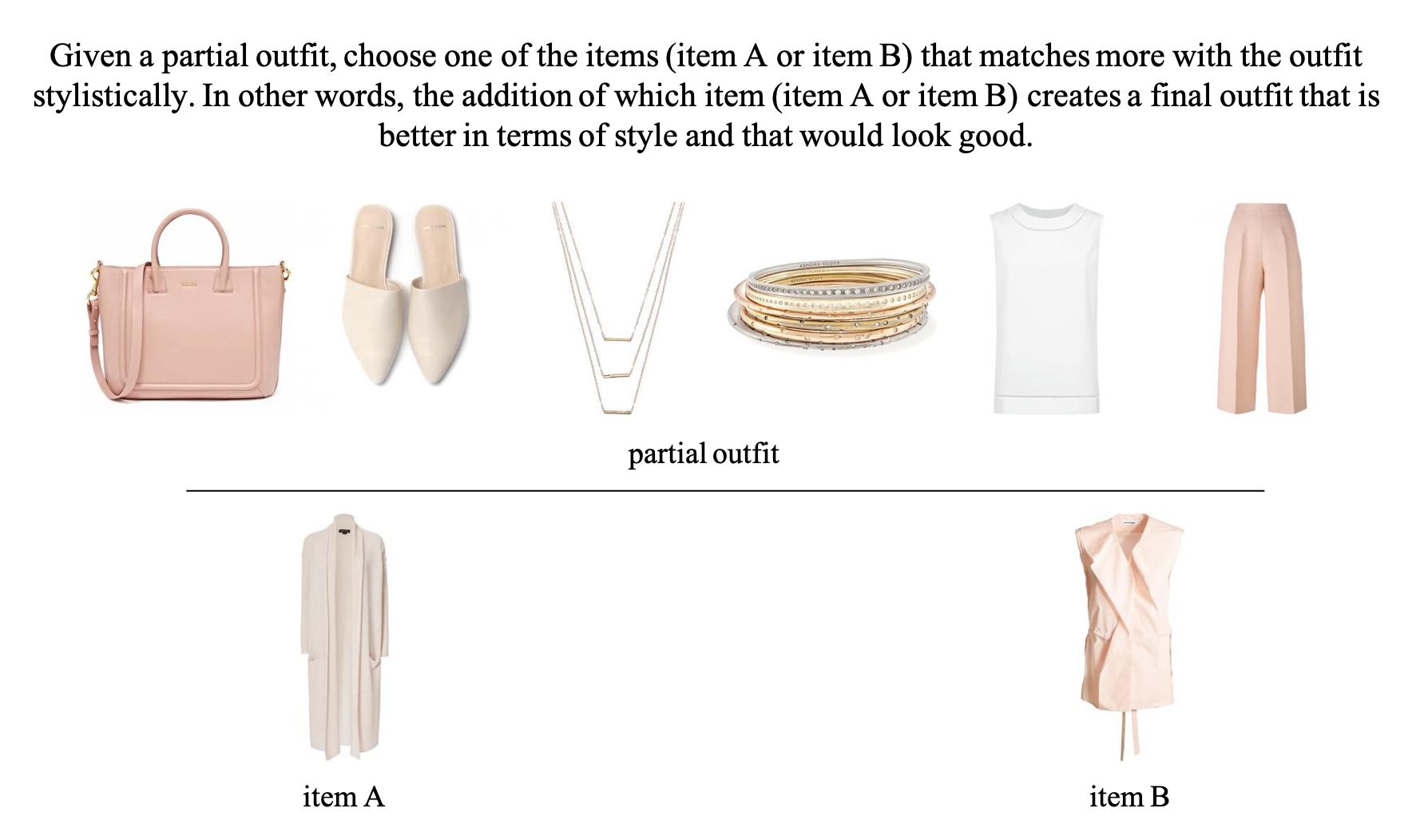}
   \includegraphics[width=\textwidth]{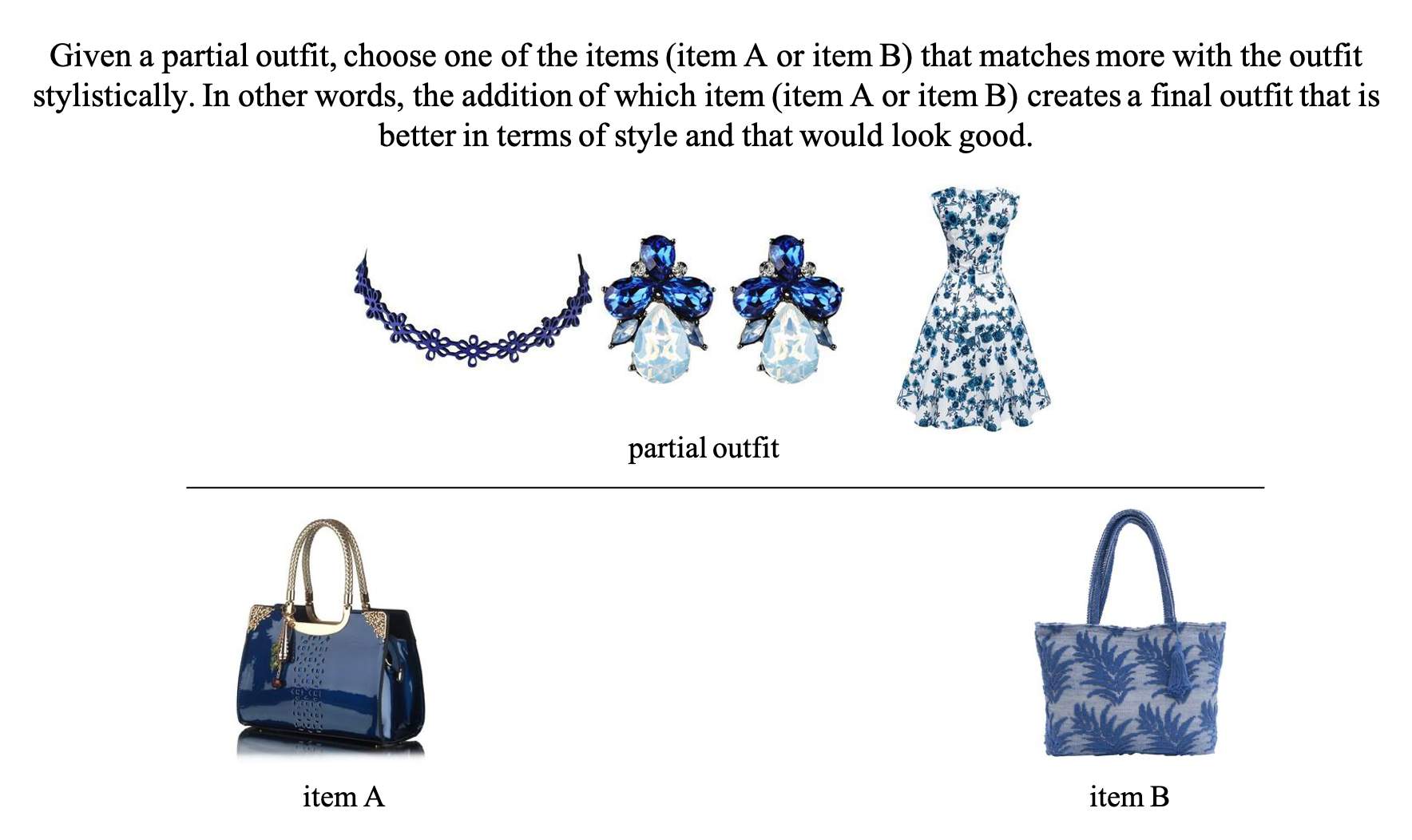}
   \end{minipage}
    \caption{Some examples of the questions presented to the MTurk annotators for the user study.  }
    \label{fig:UserStudy}
\end{figure*}

\bibliographystyle{ieee_fullname}
\bibliography{main}

\end{document}